\documentclass[journal,10pt,doublecolumn,singlespace]{IEEEtran}

\usepackage{cite,graphicx,multirow,array}

\usepackage{amsfonts,amssymb}
\usepackage[cmex10]{amsmath}
\usepackage{algorithmic}
\usepackage[ruled]{algorithm2e}	
\usepackage{epstopdf}
\usepackage{hyperref}
\usepackage{subfloat}
\usepackage{subfigure}
\usepackage{verbatim}

\usepackage[usenames,dvipsnames]{color}
\usepackage{soul}

\usepackage{colortbl}
\definecolor{mygray}{gray}{.9}
\usepackage[flushleft]{threeparttable}
\begin{document}
\title{Scale-Consistent Fusion: from Heterogeneous Local Sampling to Global Immersive Rendering}
\author{Wenpeng Xing, Jie Chen, Zaifeng Yang and Qiang Wang
	
\thanks{W. Xing, J. Chen (correspondence author) and Q. Wang are with the Department of Computer Science, Hong Kong Baptist University, Hong Kong. E-mails: \{cswpxing, chenjie and qiangwang\}@comp.hkbu.edu.hk. Z.~Yang is with the Department of Electronics and Photonics, Institute of High Performance Computing, A*STAR, Singapore. E-mail: yang\_zaifeng@ihpc.a-star.edu.sg.}}

\markboth{}%
{\MakeLowercase{\textit{Xing et al.}}:}

\maketitle

\begin{abstract}

Image-based geometric modeling and novel view synthesis based on sparse, large-baseline samplings are challenging but important tasks for emerging multimedia applications such as virtual reality and immersive telepresence. Existing methods fail to produce satisfactory results due to the limitation on inferring reliable depth information over such challenging reference conditions.
With the popularization of commercial light field (LF) cameras, capturing LF images (LFIs) is as convenient as taking regular photos, and geometry information can be reliably inferred. This inspires us to use a sparse set of LF captures to render high-quality novel views globally. However, fusion of LF captures from multiple angles is challenging due to the scale inconsistency caused by various capture settings. To overcome this challenge, we propose a novel scale-consistent volume rescaling algorithm that robustly aligns the disparity probability volumes (DPV) among different captures for scale-consistent global geometry fusion. Based on the fused DPV projected to the target camera frustum, novel learning-based modules have been proposed (i.e., the attention-guided multi-scale residual fusion module, and the disparity field guided deep re-regularization module) which comprehensively regularize noisy observations from heterogeneous captures for high-quality rendering of novel LFIs. Both quantitative and qualitative experiments over the Stanford Lytro Multi-view LF dataset show that the proposed method outperforms state-of-the-art methods significantly under different experiment settings for disparity inference and LF synthesis.

\end{abstract}

\section{Introduction}

\IEEEPARstart{R}{eproducing} photorealistic appearance of visual contents is one of the core tasks for computer graphics and computer vision.
Existing approaches fall into two major categories: physically based rendering (PBR) and image-based rendering (IBR). PBR focuses on faithful modeling of light propagation and its interactions with the environment. IBR, in contrast, works directly on images captured under real settings and renders novel views based on estimated geometry with operations such as warping and blending \cite{debevec1998efficient, zhou2018stereo}. IBR is computationally much more efficient than PBR, but its quality depends heavily on the source view's sampling pattern and the reconstruction algorithm.
In this work, we will focus on a novel globally sparse, but locally dense sampling and fusion mechanism for high-quality and scale-consistent IBR.

Recent years have witnessed blooming studies on IBR which apply deep learning techniques to boost the performance in both geometry inference and novel view synthesis. Various methods have been developed that model the 3D scene contents with different forms of representations, i.e., layered depth images \cite{ldi}, multi-plane images (MPI) \cite{zhou2018stereo, mildenhall2019llff}, point clouds \cite{10.1145/1103900.1103907}, and voxels \cite{Dai_2020_CVPR}. These representations show limitations on synthesizing complex scenes with arbitrary sampling patterns.
Neural rendering methods avoid explicit modeling of the scene geometry and directly synthesize pixels with a generative network, e.g., the DeepVoxels \cite{sitzmann2019deepvoxels} and the neural radiance fields \cite{mildenhall2020nerf}. These models require dense sampling of a compact target area and are not scalable for other large scenes.
Multi-view stereo systems \cite{yao2018mvsnet, yao2019recurrent} also work poorly over sparse inputs as they rely on continuous cost volumes for efficient geometry inference, which is both theoretically and computationally challenging with sparse source inputs.

Image-based geometric modeling and view synthesis based on \textit{sparse}, \textit{large-baseline} samplings are challenging but important tasks for emerging multimedia applications such as virtual reality and immersive telepresence. Existing methods fail to produce satisfactory results due to the limitation on  inferring  reliable  depth  information  over  such  challenging reference conditions. With the popularization of commercial light field (LF) cameras like Lytro \cite{ng2005light} and Raytrix \cite{Perwa2012Single}, capturing LF images is as convenient as taking regular photos, and geometry information can be reliably inferred.
With locally dense sampling of the target scene, reliable disparity could be estimated.
This inspires us to use a sparse set of LF captures to globally render high-quality novel views, which not only reduces the requirement for dense global sampling to a sparse set of locally dense angular sampling by the LF, but also enables us to deal with more dynamic scenes as only a few shots are required.

Fusion of LF captures from multiple angles is challenging due to the scale inconsistency caused by varying focal settings between different captures.
To overcome this challenge, we propose a novel Scale-Consistent Volume Rescaling (SCVR) algorithm which robustly aligns the disparity planes of Disparity Probability Volumes (DPV) among views for a consistent fusion. This rescaling process enables natural adaptation to various camera configurations and fuses the heterogeneous samplings into a globally consistent geometry embedding.
Based on the fused DPV projected to a target camera view, a disparity map is synthesized and refined with the target RGB capture as a guide. The LF at the novel view is subsequently synthesized and deeply regularized based on the disparity field. The synthesized LF is aligned to the direction and capture settings of the target view (including focal length, exposure, resolution, etc.). We can consequently achieve robust rendering for any heterogeneous target imagery, achieving our goal of global immersive rendering.
Both quantitative and qualitative experiments over the Stanford Lytro Multi-view LF dataset show that the proposed method outperforms state-of-the-art methods under different experiment settings, for both disparity inference and LF Synthesis.
The contributions of this paper can be generalized as:
\begin{itemize}
\item We have proposed an integrated immersive content capture scheme which can transfer the requirement of globally-dense sampling to a sparse set of locally-dense sampling (in the forms of distributed LF captures), which facilitates \textit{cheap} and \textit{convenient} capture of target scenes, especially the more dynamic ones.
\item We have proposed a novel scale-consistent frustum volume rescaling algorithm which enables fusion of \textit{distributed, heterogeneous} geometry embedding to be globally consistent.
\item We have proposed novel learning-based processing modules, i.e., attention-guided multi-scale residual fusion and deep re-regularization over the disparity field-guided LF rendering, which comprehensively regularize noisy observations from \textit{heterogeneous} captures, and fuse these complementary features for high-quality rendering of both disparity maps and novel Light Field Images (LFIs).
\item To the best of our knowledge, we have proposed the first reconstruction mechanism that can globally render a novel LF with large angular ranges and flexible optical properties (e.g., focal, exposure, and resolution settings, etc., which are determined by the target RGB image). Such a fully flexible scheme can be extended to a wide range of multi-modality sensor fusion scenarios.
\end{itemize}

The paper is organized as follows: Sec. \ref{relatedwork} introduces related works and generalizes the advantages and limitations of existing IBR representation schemes. Sec. \ref{Multi-captures Consistent Re-scaling} describes the SCVR algorithm. Sec. \ref{fvprocessing} introduces the Frustum Voxel Filtering and Attention-Guided Multi-scale Residual Fusion Module. Sec. \ref{lightfieldsynthesissection} explains the details for the disparity field synthesis and LF rendering. Section \ref{experiment} evaluates our proposed model and compare with existing methods. Sec. \ref{concludingremarks} concludes the paper.

\begin{table*}[h]
\centering
\begin{threeparttable}
\caption{Comparison of state-of-the-art geometry/appearance representation frameworks for novel view synthesis and immersive rendering. While other frameworks require specific or optimized sampling patterns and show various limitations in rendering capabilities, our method is robust to large baseline sparse inputs with diverse camera angles.}
\label{pipeline_comparison}
\begin{tabular}{|>{\arraybackslash}m{2.3cm}|>{\centering\arraybackslash}m{2.6cm}|>{\centering\arraybackslash}m{1.35cm}|>{\centering\arraybackslash}m{1.6cm}|>{\centering\arraybackslash}m{1.8cm}|>{\centering\arraybackslash}m{2.0cm}|>{\centering\arraybackslash}m{1.35cm}|>{\centering\arraybackslash}m{1.4cm}|>{\centering\arraybackslash}m{1.4cm}|}
\hline
\multirow{2}{*}{\textbf{Methods}}    &\multicolumn{2}{c|}{\textbf{Representation Scheme}}       &\multicolumn{3}{c|}{\textbf{Sampling Requirements}}   & \multicolumn{2}{c|}{\textbf{Rendering Capabilities}} \\ \cline{2-8}

&Embedding Repres. &Dimension of Repres.  & Capture Dist. (Baseline) & Camera Angle Range & Sampling Pattern &  Spatial Range &  Angle Range \\
\hline
\rowcolor{mygray} SynSin \cite{wiles2020synsin} & Feature Point Clouds & 3 & Single View & Single View  & Single View Image & Small & Small \\ \hline

MPNeuPts \cite{Dai_2020_CVPR} & Multi-plane Projected Neural Point Clouds & 3 & Large & Large  & Complete Point Clouds & Large & Large \\ \hline
\rowcolor{mygray}LBVS \cite{LearningViewSynthesis} & Disparity Map & 2 & Aperture $^\ddagger$ & Aperture $^\ddagger$ & Sparse Sub-Aperture $^\ddagger$ & Aperture $^\ddagger$ & Aperture $^\ddagger$  \\ \hline

ExtremeView \cite{extremeview} & Depth Plane Frustum Voxels (Prob. Vol.) & 2.5 $^\dagger$ & Very Small & Very Small & Locally Sparse & Small & Large \\ \hline

\rowcolor{mygray}FlexLF \cite{9204825}  & Disparity Map  & 2 & Aperture $^\ddagger$  & Aperture $^\ddagger$   & Sparse Sub-Aperture $^\ddagger$ & Aperture $^\ddagger$ & Aperture $^\ddagger$ \\ \hline

LLFF \cite{mildenhall2019llff} & Multiple Multi-Plane Images & 2.5 & Px/DP $^\mathsection$ & Small (Ideally Fronto-Parallel) & Irregular Local Grid & Small & Small\\ \hline

\rowcolor{mygray} MVSNet \cite{yao2018mvsnet} & Depth Plane Frustum Voxels (Cost Vol.) & 2.5 $^\dagger$ & Small & Small (Ideally Fronto-Parallel) & Globally Dense & Small & Small \\\hline

DeepVoxels \cite{sitzmann2019deepvoxels} &  Feature Voxels & 3 & Small &  Small & Globally Dense (Hemisphere) & Large & Large\\ \hline

\rowcolor{mygray}NeRF \cite{mildenhall2020nerf} & Multi-layer Perceptron Parameters & 5 & Small & Small  & Globally Dense (Hemisphere) & Large & Large  \\ \hline

Ours  & Depth Plane Frustum Voxels (Prob. Vol.) & 2.5 $^\dagger$ & Large & Large & Globally Sparse & Large & Large   \\ \hline

\end{tabular}
\begin{tablenotes}
  \small
  \item $\dagger$ \ul{\textit{2.5}} stands for the dimension for representations in which the z dimension is coarsely sampled (e.g., by depth planes).
  \item $\ddagger$ \ul{\textit{Aperture}} indicates that the distance and angle differences between two adjacent captures are within the camera main lens' aperture diameter and its field-of-view angle, respectively.
  \item $\mathsection$ \ul{\textit{Px/DP}} indicates maximum camera lateral movement which causes shift of \ul{\textit{one pixel}} between any adjacent depth planes \cite{Srinivasan_2019_CVPR}.
\end{tablenotes}
\vspace{-0.4cm}
\end{threeparttable}
\end{table*}

\section{Related work} \label{relatedwork}

For IBR, one of the most critical assumptions is accurate depth estimation. In this section, we will first review related works on depth inference based on different imaging modalities and sampling conditions. Then, different geometry embedding representations and novel view rendering methods will be reviewed and analyzed.

\subsection{Scene Depth Inference}
\textit{\textbf{Depth from Multi-View Stereo (MVS).}}
The problem of reconstructing the geometry from multi-view images is known as Structure from Motion (SfM) \cite{hartley_zisserman_2004}, which starts with local feature extraction, matching, geometric verification, followed by image registration, triangulation and bundle adjustment - which filters out outliers for a refined depth reconstruction.
Newcombe et al.\cite{newcombe2011dtam} proposed a real-time camera tracking and scene reconstruction system by minimizing a global regularized energy function in a non-convex optimization framework. Pizzoli et al. \cite{pizzoli2014remode} estimated the depths of pixels by searching for their correspondence points from multi-view images, and updates the point position by a robust probabilistic model. 
Schoenberger et al.\cite{schoenberger2016sfm} presented a SfM system, named as COLMAP, to reconstruct the scenes by leveraging pixel-wise photometric and geometric priors.

Conventional SfM/MVS algorithms rely on the photometric consistency assumption and show limitations in handling textureless regions and reflective surfaces.
Taking advantage of the feature descriptive power of the Convolutional Neural Networks (CNN), recent works use CNNs to extract discriminative features that encode local and global information as similarity measures and have achieved significant improvements. In particular, Yao et al. \cite{yao2018mvsnet} proposed the MVSNet which warps the reference images' feature volumes to the canonical frustum and calculates the feature matching cost which is subsequently regularized by a multi-scale 3D-UNet for the prediction of depth. Several methods have been proposed to improve the MVSNet by reducing memory cost and computational time, such as modifying the overall pipeline to a coarse-to-fine \cite{Yang_2020_CVPR}, sequential structure \cite{yao2019recurrent}, adding additional edge convolution \cite{ChenPMVSNet2019ICCV} or Gauss-Newton layer \cite{Yu_2020_fastmvsnet} to recover better scene geometric details. This line of methods, generalized as the entry MVSNet in TABLE \ref{pipeline_comparison}, rely on building a cost volume with continuous angular variation for efficient inference of correspondence among the depth planes, which show serious limitation dealing with sparse, large baseline reference inputs.

\textit{\textbf{Depth from Light Field Images.}}
Estimating depth from LFIs is essentially a multi-view stereo problem with denser and more regular angular sampling of the scene. Depth maps can be estimated by analyzing the slopes of lines in the epipolar-plane images (EPI) \cite{2012Globally, 2013Scene}.
In addition to EPI clues, depth can be estimated from the defocus and correspondence clues \cite{2016Occlusion}. Robust regularizers such as super-pixels \cite{chen2018accurate} and the occlusion-aware indicators \cite{2015Accurate} can be applied to improve the boundary accuracy.
In the spectrum of applying deep learning techniques for the task, Heber et al. \cite{Heber_2017_ICCV} applied 3D convolutions to EPI volumes in a U-shape network architecture.
Peng et al. \cite{peng2018unsupervised} designed a combined loss function imposing both compliance and divergence constraints on the warped SAIs to the central view to predict the disparity in an unsupervised framework.
Shin et al. \cite{Shin_2018_CVPR} utilized a multi-stream input including SAIs organized in horizontal, vertical, left, and right diagonal directions to fully explore the angle information.
Guo et al.\cite{guo2020acurate} proposed to train a sub-network to explicitly predict the occlusion regions for better handling of these most challenging areas. These prediction methods, represented by the entries LBVS and FLexLF in TABLE \ref{pipeline_comparison}, can only handle small spatial/angular baseline inputs within the camera aperture.

\textit{\textbf{Depth from Monocular Image.}}
Estimating scene depth based on a single image is challenging as it lacks reliable geometrical clues from multi-view observations.
With the rise of deep learning techniques that learn and consolidate clues such as shading, texture, and semantics, models trained over RGB-D datasets\cite{menze2015object} demonstrate visually and perceptually satisfying depth prediction results. Ren et al.\cite{Ranftl2020Towards} proposed a zero-shot cross-dataset training strategy to leverage a variety of datasets. Nonetheless, the prediction accuracy based on monocular images cannot be guaranteed due to the scale and semantic ambiguities.

\subsection{Scene Representation frameworks for Novel View Synthesis}
As generalized in TABLE \ref{pipeline_comparison}, several seminal scene geometry/appearance representation frameworks have been developed in recent years, which are able to synthesize high-quality views from novel angles. We review these representations and analyze their respective advantages and limitations.

\textit{\textbf{Multi-plane Images.}}
The seminal multi-plane image (MPI) model \cite{zhou2018stereo, mildenhall2019llff, Srinivasan_2019_CVPR} analyzes the scene geometry over plane-sweep volumes and projects pixels into a set of fronto-parallel RGB-$\alpha$ planes within the reference camera’s view frustum. Novel views are rendered by homography-warping and over compositing the RGB-$\alpha$ planes. Based on the MPI representation, Flynn et al. \cite{Flynn_2019_CVPR} incorporated learned gradient descent mechanism. Srinivasan et al. \cite{Srinivasan_2019_CVPR} enforced a layer constraint for the occluded pixels. Choi et al. \cite{extremeview} dealt with occlusion and depth uncertainty with a depth probability volume and combined multiple warping clues over large extrapolation angles for robust fusion. Local Light Field Fusion \cite{mildenhall2019llff} renders novel views by blending re-weighted MPIs generated from a set of input images. This method requires capture locations to be approximately on the same plane so that the discretized depth planes among views could be aligned with each other for efficient fusion.
Represented by the entries LLFF and MVSNet in TABLE \ref{pipeline_comparison}, methods based on \textit{layered homography transform} assume a flattened fronto-parallel depth distribution with only rotating or planar-moving camera motions, which limit its performance in synthesizing complex scenes over flexible camera shooting angles. In addition, the maximum camera lateral movement is limited by its causal shift between any adjacent depth plane by one pixel \cite{Srinivasan_2019_CVPR}.

\textit{\textbf{Neural Implicit Functions.}}
As an alternative to conventional IBR methods, neural rendering methods avoid explicit warping operations and directly encode the scene into the parameters of an Multi-layer Perception (MLP) network which can map spatial 3D query points into RGB-$\alpha$ values \cite{mildenhall2020nerf}, signed distance values \cite{Park_2019_CVPR, Jiang_2020_CVPR} or occupancy field values \cite{Genova_2020_CVPR}.
Sitzmann et al. \cite{sitzmann2019deepvoxels} proposed to learn a DeepVoxels representation from 2D images without explicitly modeling the scene geometry. Mildenhall et al. \cite{mildenhall2020nerf} proposed to represent the scene as a continuous volumetric function which directly maps a 5D coordinate (for the observation ray) to output colors and densities. As represented by the entries DeepVoxels and NeRF in TABLE \ref{pipeline_comparison}, these models require dense sampling (more than 100 captures) of a compact target area and are not scalable for larger scenes.

\begin{figure*}[ht]{
\centering
\includegraphics[width=1\linewidth]{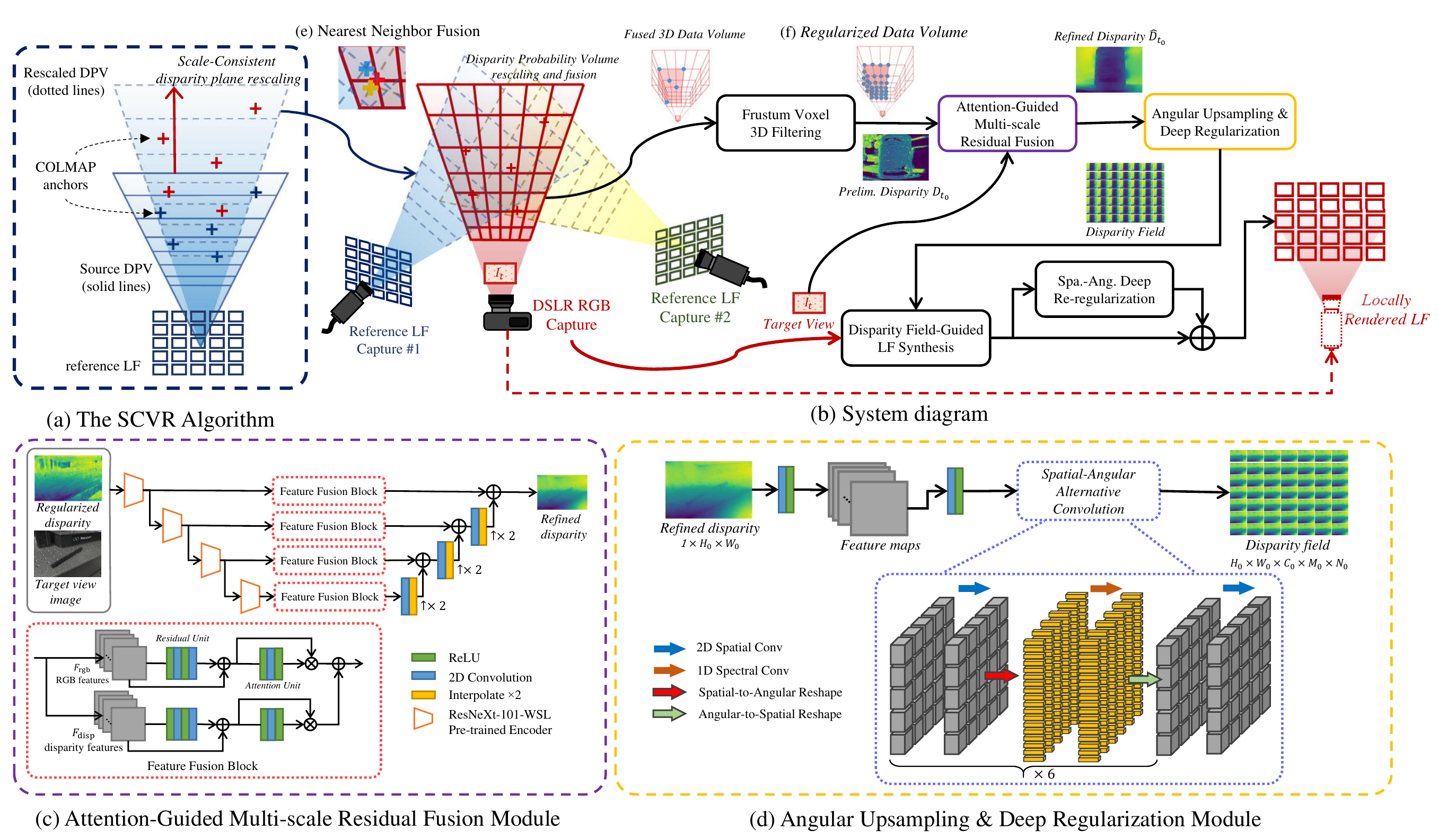}
\caption{The overall pipeline of our proposed method. 1) The source \textit{heterogeneous} LFs' DPVs are rescaled to be scale-consistent by our Scale-Consistent Volume Rescaling algorithm, then the disparity planes of the DPVs are aligned for accurate warping and blending. 2) A Frustum Voxel Filtering Module is proposed to complete the DPVs. 3) The Attention-Guided Multi-scale Residual Fusion Module performs RGB-D fusion to produce a high-quality disparity map. 4) The LF is finally synthesized by backward warping pixels of target view image with estimated accurate disparity prior and further refined by spatial-angular convolutional blocks. }\label{fig:overall}\vspace{-0.4cm}
}\end{figure*}

\textit{\textbf{Point-based Representations.}}
Using points as rendering primitives has been an active research topic recently \cite{KOBBELT2004801}. Point clouds bear explicit geometric properties which make them flexible to rotate to an arbitrary viewpoint for rendering. Points can be encoded as latent appearance vectors and rendered by a deep frame buffer for realistic effects \cite{Meshry_2019_CVPR}. Wiles et al. \cite{wiles2020synsin} used spatial feature network to extract higher-level representation from input RGB colors. Feature point clouds are then transformed and rendered by the neural point cloud renderer, which softens the hard z-buffer via using alpha over-compositing. The unstructured nature of point clouds has driven most researchers to transform them to regular 3D voxel grids. For example, multi-plane based voxelization and multi-plane rendering are adopted in \cite{Dai_2020_CVPR}.
As represented by the entry \textit{SynSin} and \textit{MPNeuPts} in TABLE \ref{pipeline_comparison}, rendering the unstructured point clouds requires computationally expensive pre-processing (e.g., voxelization) and z-buffering operations.

\textit{\textbf{Local Immersive Light Field Synthesis.}}
LFI itself is also a powerful scene representation with scene geometry and view-dependent effects embedded within its 4D spatial-angular structures.
Researchers have focused on synthesizing complete LFIs from sparse inputs.
Early algorithms rely on the plenoptic function and take LF synthesis as a re-sampling task \cite{4032816}. Without accurate geometry estimations, this approach can only synthesize novel views with small baselines.
Some works attempt to synthesize LFIs with geometric \cite{1238625}.
Some enforce priors over the Fourier spectrum \cite{10.1145/2682631} and the Shearlet transform domain \cite{7817742}.
Kalantari et al.\cite{LearningViewSynthesis} made the first attempt to use CNNs to synthesize novel LFIs from a sparse set of inputs, with a system that can be broken down into disparity and color estimation modules. Srinivasan et al.\cite{srinivasan2017learning} used two convolutional neural networks to synthesize a 4D RGB-D LF from a 2D RGB image, where the first CNN estimates the scene geometry and the second predicts occluded rays and non-Lambertian effects.
Yeung et al.\cite{yeung2018fast} introduced a fast LF reconstruction from a sparsely-sampled LF in a coarse-to-fine manner and explored the dense spatial and angular clues by spatial-angular alternative convolutions. Wu et al. \cite{8099661} restored the angular detail by using Deep Convolutional Network on EPI.

\section{Proposed Method}

Given $N_\text{src}$ sets of source LFIs $\{\mathbf{L}_t\in \mathbb{R}^{H_t\times W_t\times M_t\times N_t}\}_{t=1}^{N_\text{src}}$ which capture the scene from \textit{large diverse} viewing positions, our model $\mathcal{F}(\cdot)$ aims to accurately fuse these source information and synthesize a novel immersive LF $\mathbf{\hat{L}}_{t_0}\in \mathbb{R}^{H_{0}\times W_{0}\times C_0 \times M_{0}\times N_{0}}$ which is aligned to an image capture $I_{t_0}\in\mathbb{R}^{H_0\times W_0\times C_0}$ at a target angle $t_0$:
\begin{equation}
\mathbf{\hat{L}}_{t_0}=\mathcal{F}(\{\mathbf{L}_t\}_{t=1}^{N_\text{src}}, I_{t_0}; \theta),
\end{equation}
here $H_t, W_t$ indicate the spatial dimensions and $M_t, N_t$ indicate the angular dimensions of the source LFIs, respectively. $H_0, W_0, C_0$ indicate the spatial and channel dimensions of the target RGB capture, and $M_0, N_0$ indicate the angular dimensions of the synthesized immersive LFI $\mathbf{\hat{L}}_{t_0}$. $\theta$ stand for the parameters of the learning modules.

The overall pipeline of our proposed framework is illustrated in Fig.~\ref{fig:overall}, which consists of five steps.
\textit{\ul{First}}, a Disparity Probability Volume\footnote{DPV is calculated with the slope of EPI line which is proportional to the disparity value \cite{chen2018accurate}, the confidence of the slope estimation represents the probability of its corresponding disparity value.} (DPV) $\{\mathbf{V}_t\in \mathbb{R}^{W_t\times H_t\times N_\text{dp}}\}_{t=1}^{N_\text{src}}$ is estimated for each of the source LFI. Here $N_\text{dp}$ stands for the number of disparity/depth planes. To ensure all the DPVs are scale-consistent, we rely on COLMAP \cite{schoenberger2016sfm, schoenberger2016mvs}, a well adopted structure-from-motion pipeline, to estimate the source view camera poses and establish a set of 3D points as reliable 3D anchors among the source LFIs. The Scale-Consistent Volume Rescaling (SCVR) algorithm is proposed to realign the DPVs based on these reliable 3D anchors, which produces a set of scale-consistent DPVs $\{\mathbf{U}_t\}_{t=1}^{N_\text{src}}$.
\textit{\ul{Second}}, the disparity probability values within the source rescaled DPVs $\{\mathbf{U}_t\}_{t=1}^{N_\text{src}}$ are \emph{homography warped} to the fronto-parallel planes of the target camera's frustum, and fused as a 3D data volume $\mathbf{U}_{t_0}$.
\textit{\ul{Third}}, to comprehensively exploit the geometrical information in $\mathbf{U}_{t_0}$, the Frustum Voxel Filtering Module is deployed to explore the information in $\mathbf{U}_{t_0}$ for a coarse disparity estimation $D_{t_0}\in\mathbb{R}^{H_0\times W_0}$.
\textit{\ul{Fourth}}, the Attention-Guided Multi-scale Residual Fusion Module is designed to refine $D_{t_0}$ with the textural and semantic guides from the target view image $I_{t_0}$ in a multi-scale, progressive manner, and produce the final disparity estimation $\hat{D}_{t_0}$.
\textit{\ul{Finally}}, a  Disparity Field-Guided Deep  Re-regularization Module is proposed to raise $I_{t_0}$ into a locally immersive LF $\mathbf{\hat{L}}_{t_0}$ by back-warping  and deep spatial-angular regularization. In the following subsections, we will elaborate the details for each sub-module.

\subsection{The Scale-Consistent Volume Rescaling Algorithm}\label{Multi-captures Consistent Re-scaling}

Depth estimation based on a single LF is essentially a multi-view stereo problem with the virtual cameras of SAIs \textit{densely and regularly} positioned, enabling efficient algorithms to be developed for high precision sub-pixel level parallax estimation.
A DPV $\mathbf{V}_t(x,y,d)\in\mathbb{R}^{H_t\times W_t\times N_\text{dp}}$ can be estimated, which reflects each pixel's probability distribution with respect to each disparity plane.
Based on the DPV $\mathbf{V}_t(x,y,d)$, an initial disparity map $D_\text{init}$ can be calculated \cite{2016Occlusion,chen2018accurate}.

The exact \textit{object space distance} that each DPV's \textit{disparity interval} corresponds to varies between different captures, which is decided by the capture settings (e.g., focal length settings).
Therefore, before $\{\mathbf{V}_t\}_{t=1}^{N_\text{src}}$ can be coherently fused together to the target camera's viewing frustum, a \textit{rescaling} process must be implemented to ensure the scale-consistency of the DPVs among the source views.
To this end, we propose a Scale-Consistent Volume Rescaling algorithm which iteratively performs rescaling, trimming and interpolation operations over the DPVs.

\begin{algorithm}[t]
\caption{The Scale-Consistent Volume Rescaling algorithm (SCVR).}
\SetAlgoLined
- \textbf{Input}: Sparse 3D anchors $\mathcal{P}$; DPV $\mathbf{V}_t$;  COLMAP depth range $\kappa$; Iterative times $N$; Initial disparity map $D_\text{init}$. \\ 
- \textbf{Output}: rescaled DPV $\mathbf{\bar{U}}_t$.\\
- \textbf{Algorithm}:

- Initialize $\{\alpha,\beta\}$ as Eq. (\ref{eq:calparams}) with $D_\text{init}$ and $\mathcal{P}$;

\While{$ n \leq N $}{

 Update the $\Psi_{\alpha,\beta}$ and  $\{\alpha,\beta\}$ as Eq. (\ref{eq:mapping}, \ref{eq:calparams});

Update the $\mathbf{V}_t$ with $\Psi_{\alpha,\beta}$ to produce $\mathbf{U}_t$ as Eq. (\ref{eq:rescale});

 Discard planes exceeding $\kappa$ from $\mathbf{U}_t$ to produce $\mathbf{U}^\gamma_t$ as Eq. (\ref{eq:trim});

 Interpolate the $\mathbf{U}^\gamma_t$ to $\mathbf{\bar{U}}_t$ as Eq. (\ref{eq:interp});

 $n=n+1$;
}
\end{algorithm}

We first use COLMAP \cite{schoenberger2016sfm} to analyze the scene geometry based on the central views of the source LFIs $\{I_t\}_{t=1}^{N_\text{src}}$ and the 2D image $I_{t_0}$ from the target viewing angle. COLMAP outputs the camera parameters for each view, along with a set of \textit{sparse} but \textit{reliable} 3D points $\mathcal{P}(x,y,z)$ which establish correspondences among the 2D pixels from $I_{t_0}\cup \{I_t\}_{t=1}^{N_\text{src}}$.
With slight abuse of notation, we use ${\mathcal{P}_t(x,y)}$ to represent the 2D image coordinates in $\{I_t\}_{t=1}^{N_\text{src}}$ which are projected \textit{from} the 3D points $\mathcal{P}(x,y,z)$ in \textit{world coordinate}, and use $\mathcal{P}_t(z)$ to represent the depth of $\mathcal{P}(x,y,z)$ with respect to the image plane of view $t$. The 3D point $\mathcal{P}(x,y,z)$ is visible to multiple views, the depth $\mathcal{P}_t(z)$ is a cross-view consistent value in \textit{world coordinate}. 

We want to rescale the DPVs to be cross-view consistent via minimizing the difference between depth $\mathcal{P}_t(z)$ and its corresponding disparity value in $D_\text{init}$.
Therefore, we transform the initial disparity estimation $D_\text{init}(\mathcal{P}_{t}(x,y))$ to depth values via:
\begin{equation}\label{eq:mapping}
\Psi_{\alpha,\beta}(D_\text{init}(\mathcal{P}_{t}(x,y)))=\frac{\alpha}{D_\text{init}(\mathcal{P}_{t}(x,y))}+\beta.
\end{equation}
Here $\Psi_{\alpha,\beta}$ is a mapping function that reflects the inverse-proportional relationship between the absolute depth and the capture-specific disparity.  $\alpha$ is related to the focal length and the camera baseline; $\beta$ is the bias parameter. $\{\alpha,\beta\}$ can be determined by minimizing the alignment error given by
\begin{equation}\label{eq:calparams}
\{\alpha,\beta\}=\mathop{\arg\min}_{\hat{\alpha},\hat{\beta}}||\Psi_{\hat{\alpha},\hat{\beta}}(D_\text{init}(\mathcal{P}_t(x,y)))-\mathcal{P}_t(z)||^2_2.
\end{equation}
The estimated parameters $\{\alpha,\beta\}$ can subsequently be used to transform all disparity planes in $\mathbf{V}_t$ to produce $\mathbf{U}_t$:
\begin{equation}\label{eq:rescale}
\mathbf{U}_t=\Psi_{\alpha,\beta}(\mathbf{V}_t).
\end{equation}

After each transformation iteration, depth planes in $\mathbf{U}_t$ that exceeds COLMAP's depth range $\kappa$ will be trimmed and discarded:
\begin{equation}\label{eq:trim}
\mathbf{U}^\gamma_t=\Gamma_\text{trim}(\mathbf{U}_t): \mathbb{R}^{H\times W\times N_\text{dp}}\mapsto\mathbb{R}^{H\times W\times N'_\text{dp}},
\end{equation}
where $N_\text{dp}^{'}$ is the remaining number of planes in $\mathbf{U}^\gamma$ after trimming.
This is subsequently followed by a linear interpolation process along the depth dimensions $N'_\text{dp}$ to produce a new probability volume $\mathbf{\bar{U}}_t$ which preserves its original disparity plane resolution $N_{dp}$:
\begin{equation}\label{eq:interp}
\mathbf{\bar{U}}_t=\Gamma_\text{interp}(\mathbf{U}^\gamma_t): \mathbb{R}^{H\times W\times N'_\text{dp}}\mapsto\mathbb{R}^{H\times W\times  N_\text{dp}}.
\end{equation}
Note that the procedures defined from Eq.~(\ref{eq:mapping}) to Eq. (\ref{eq:interp}) will be repeated for $N$ times until convergence. In our experiments, $N$=7 generally gives satisfactory results. This iteration process helps to progressively remove redundant planes. Each iteration generates a more \textit{compact} volume with higher disparity resolution, ensuring higher alignment accuracy than the one-off rescaling.

In summary, the operator  $\Psi_{\alpha,\beta}$ \textit{stretches} or \textit{compresses} the DPVs to be scale-consistent among all source views; $\Gamma_\text{trim}$ selects the depth cut-off range; and $\Gamma_\text{interp}$ interpolates the DPVs to a consistent resolution.
The pseudo code of the SCVR algorithm has been generalized in Algorithm 1.

\textbf{Remark.} 
The SCVR algorithm is based on reliable COLMAP spatial anchors. These anchors are crucial for consistent fusion of \textit{heterogeneous} source volumes, and for flexible projection with accurate alignment to target RGB capture's optical properties.

\begin{figure}[t]{
\centering
\includegraphics[width=1\linewidth]{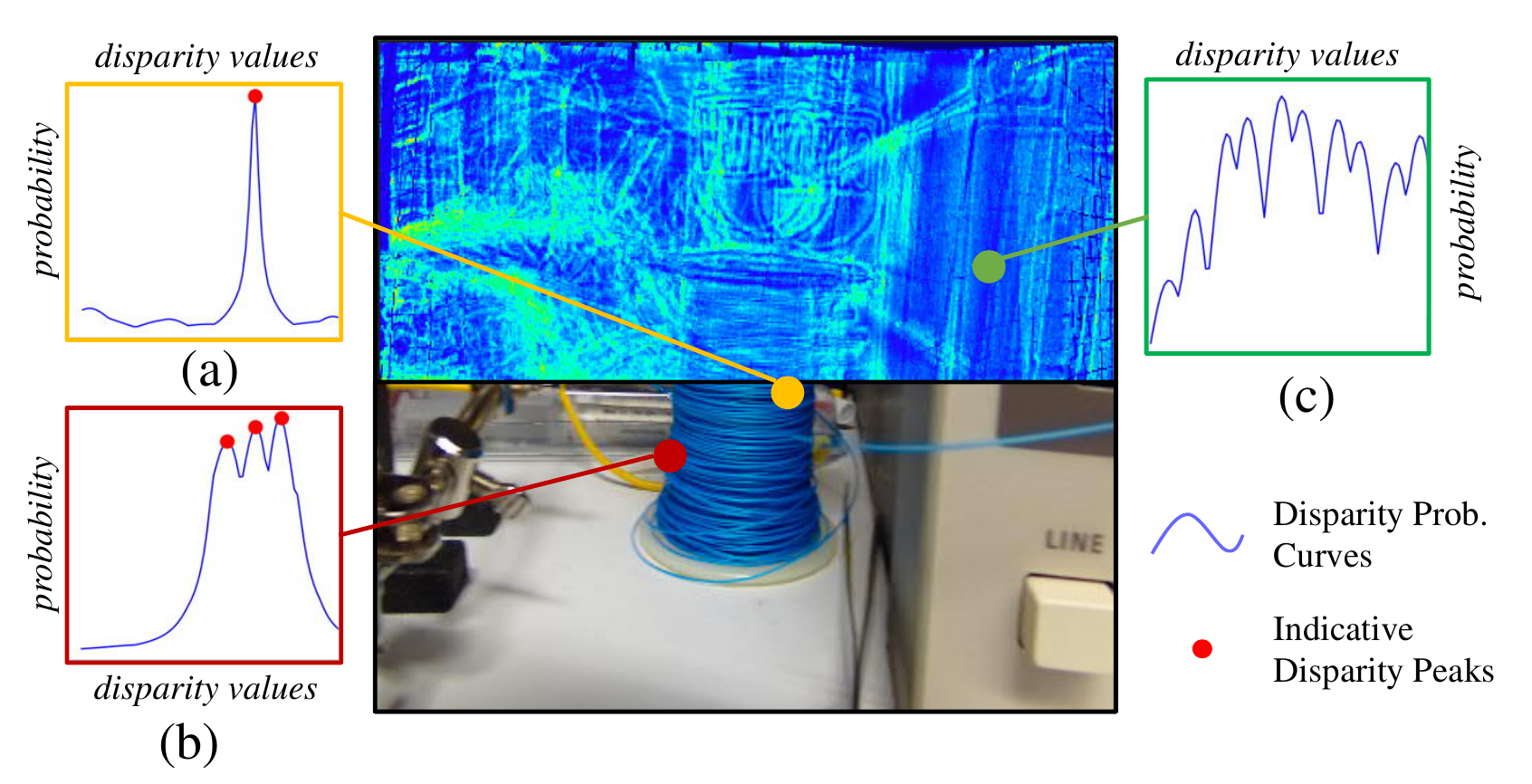}
\caption{Disparity probability distribution along a light ray.}
\label{fig:dispdist}
}\end{figure}

\begin{figure*}[h]{
\centering

\begin{subfigure}{
\centering
\begin{minipage}[t]{0.2\textwidth}
\centering
\includegraphics[width=1.3in]{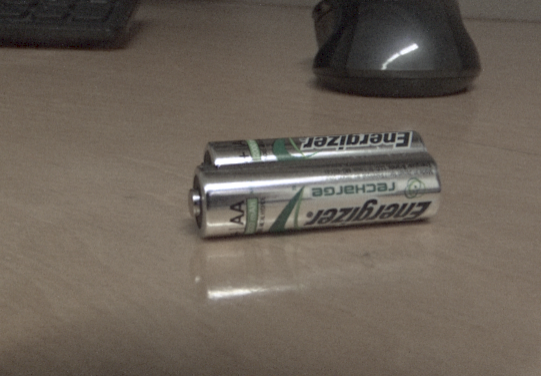}
\end{minipage}
\hspace{-0.2in}
\begin{minipage}[t]{0.2\textwidth}
\centering
\includegraphics[width=1.3in]{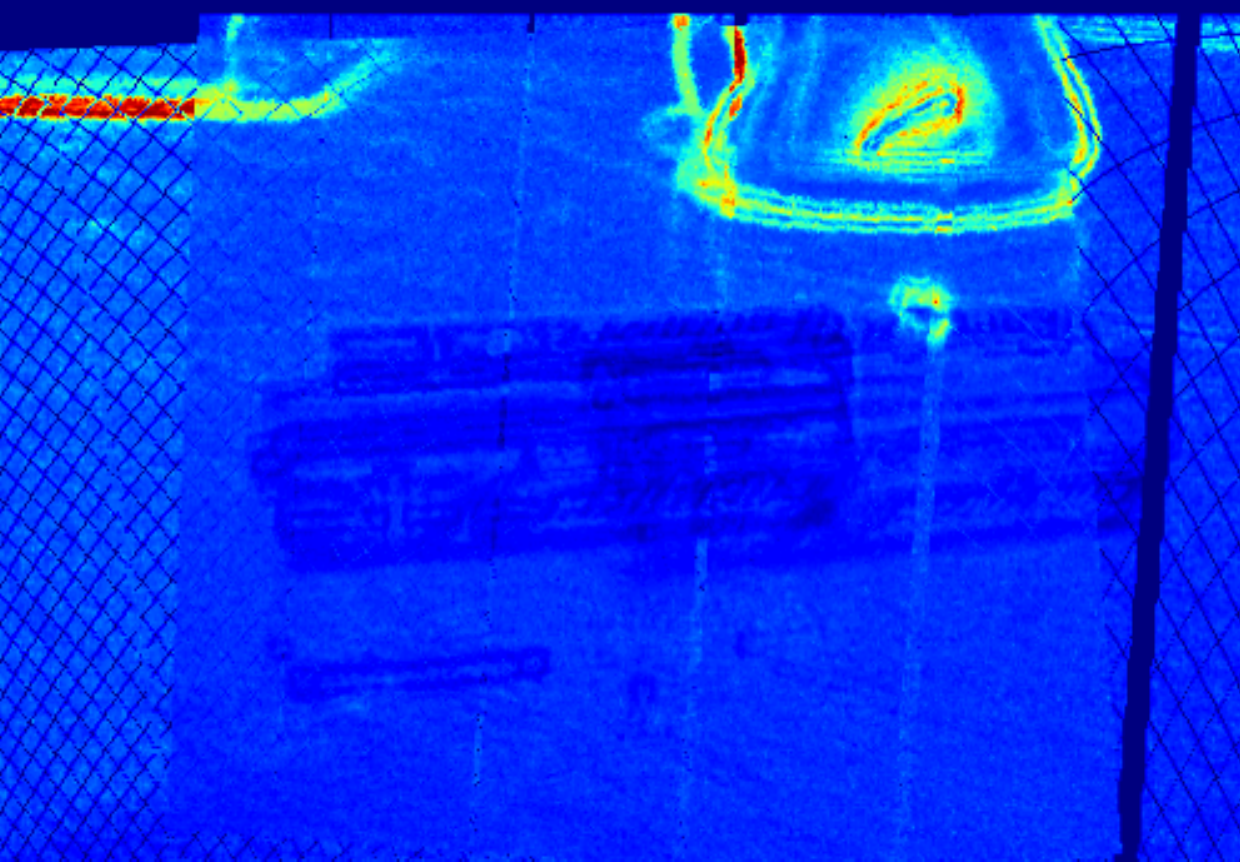}
\end{minipage}
\hspace{-0.2in}
\begin{minipage}[t]{0.2\textwidth}
\centering
\includegraphics[width=1.3in]{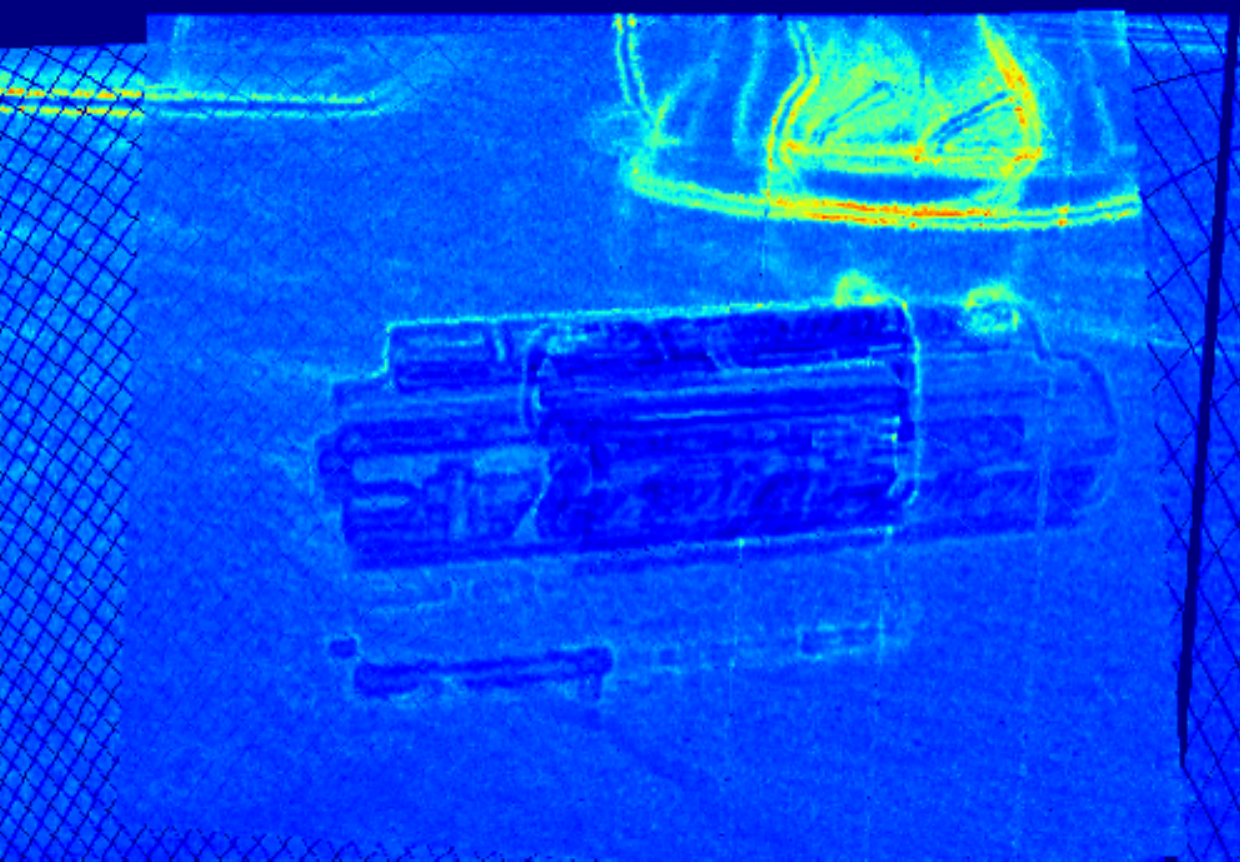}
\end{minipage}
\hspace{-0.2in}
\begin{minipage}[t]{0.2\textwidth}
\centering
\includegraphics[width=1.3in]{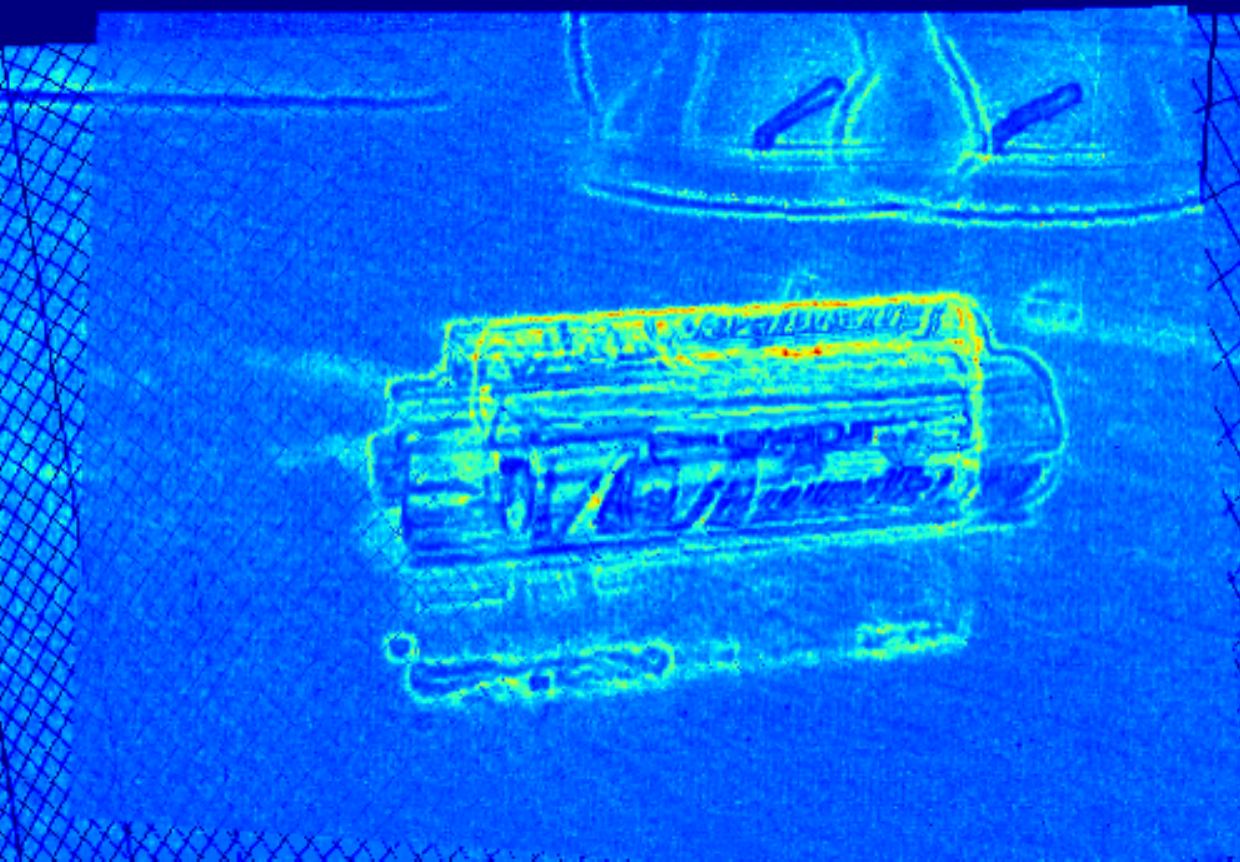}
\end{minipage}
\hspace{-0.2in}
\begin{minipage}[t]{0.2\textwidth}
\centering
\includegraphics[width=1.3in]{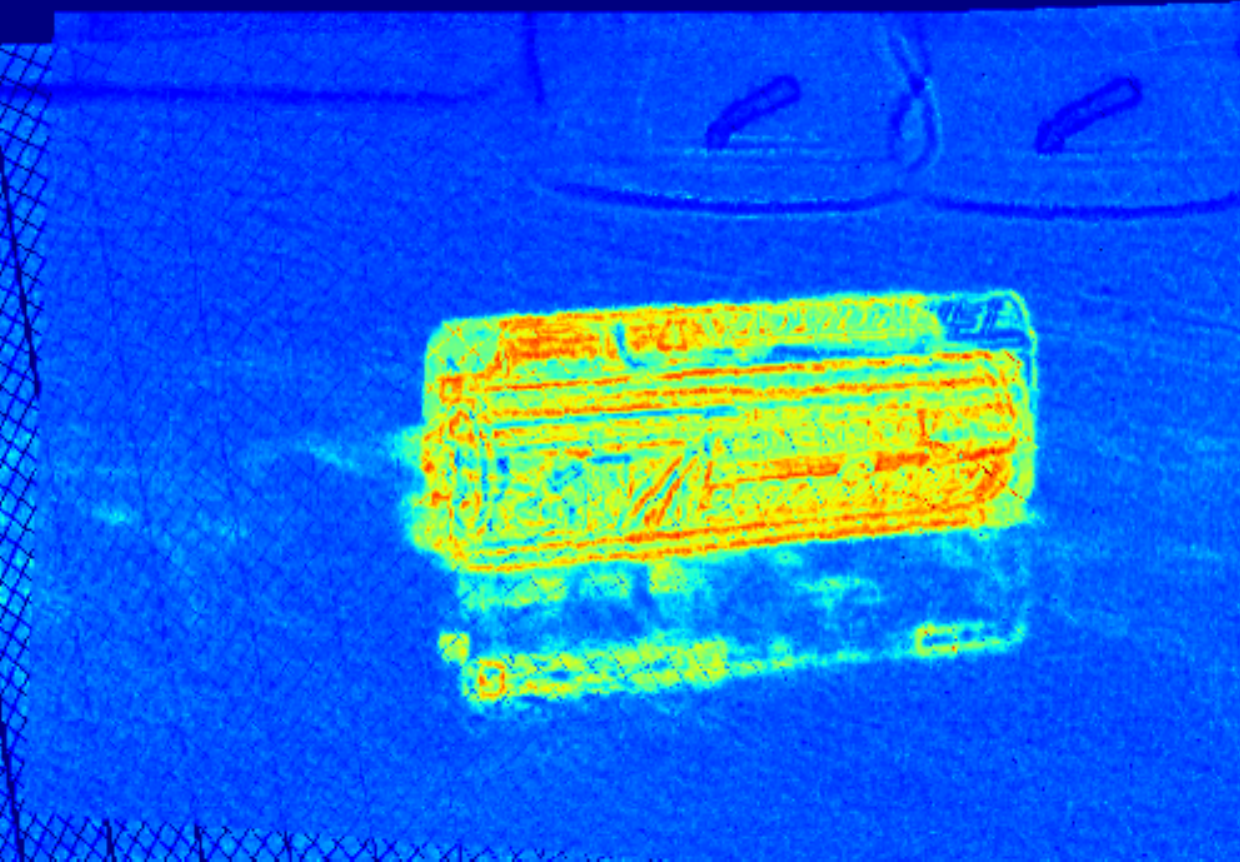}
\end{minipage}
}\end{subfigure}

\vspace{-0.12in}
\begin{subfigure}{
\centering
\begin{minipage}[t]{0.2\textwidth}
\centering
\includegraphics[width=1.3in]{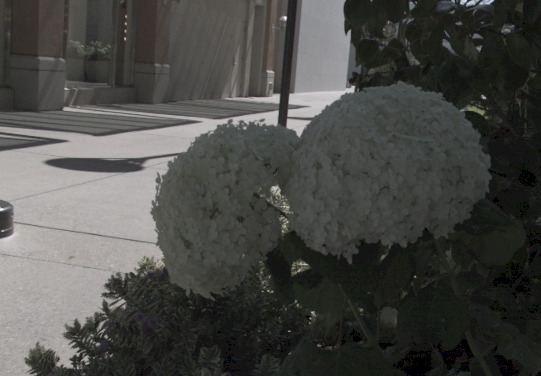}
\end{minipage}
\hspace{-0.2in}
\begin{minipage}[t]{0.2\textwidth}
\centering
\includegraphics[width=1.3in]{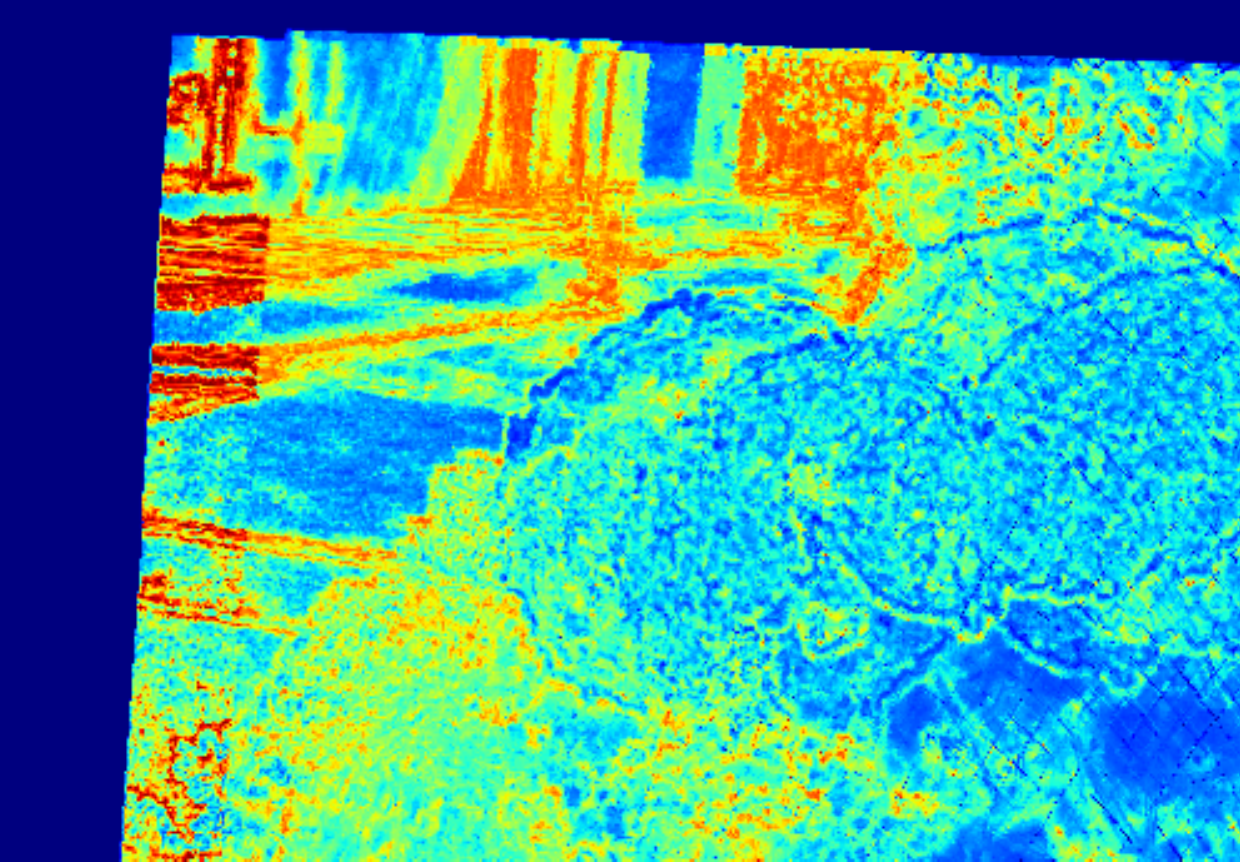}
\end{minipage}
\hspace{-0.2in}
\begin{minipage}[t]{0.2\textwidth}
\centering
\includegraphics[width=1.3in]{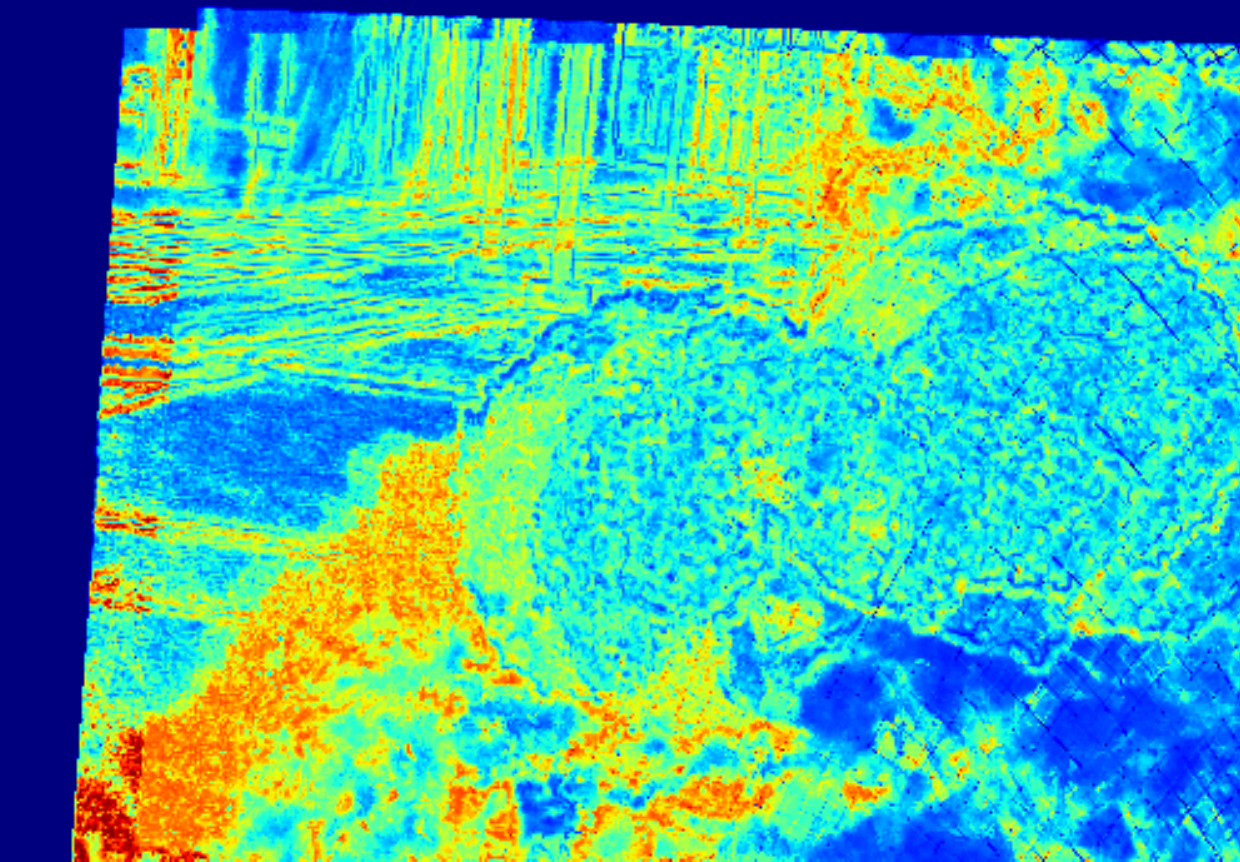}
\end{minipage}
\hspace{-0.2in}
\begin{minipage}[t]{0.2\textwidth}
\centering
\includegraphics[width=1.3in]{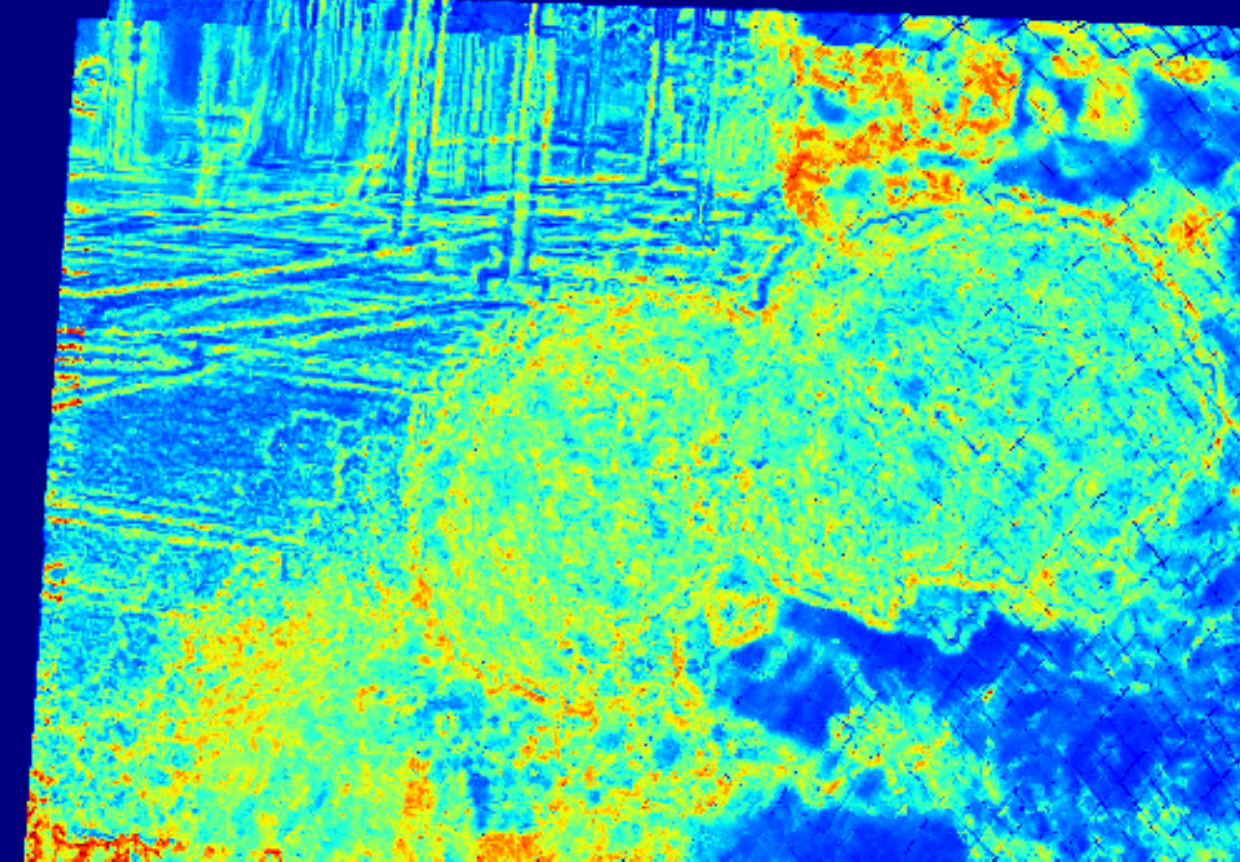}
\end{minipage}
\hspace{-0.2in}
\begin{minipage}[t]{0.2\textwidth}
\centering
\includegraphics[width=1.3in]{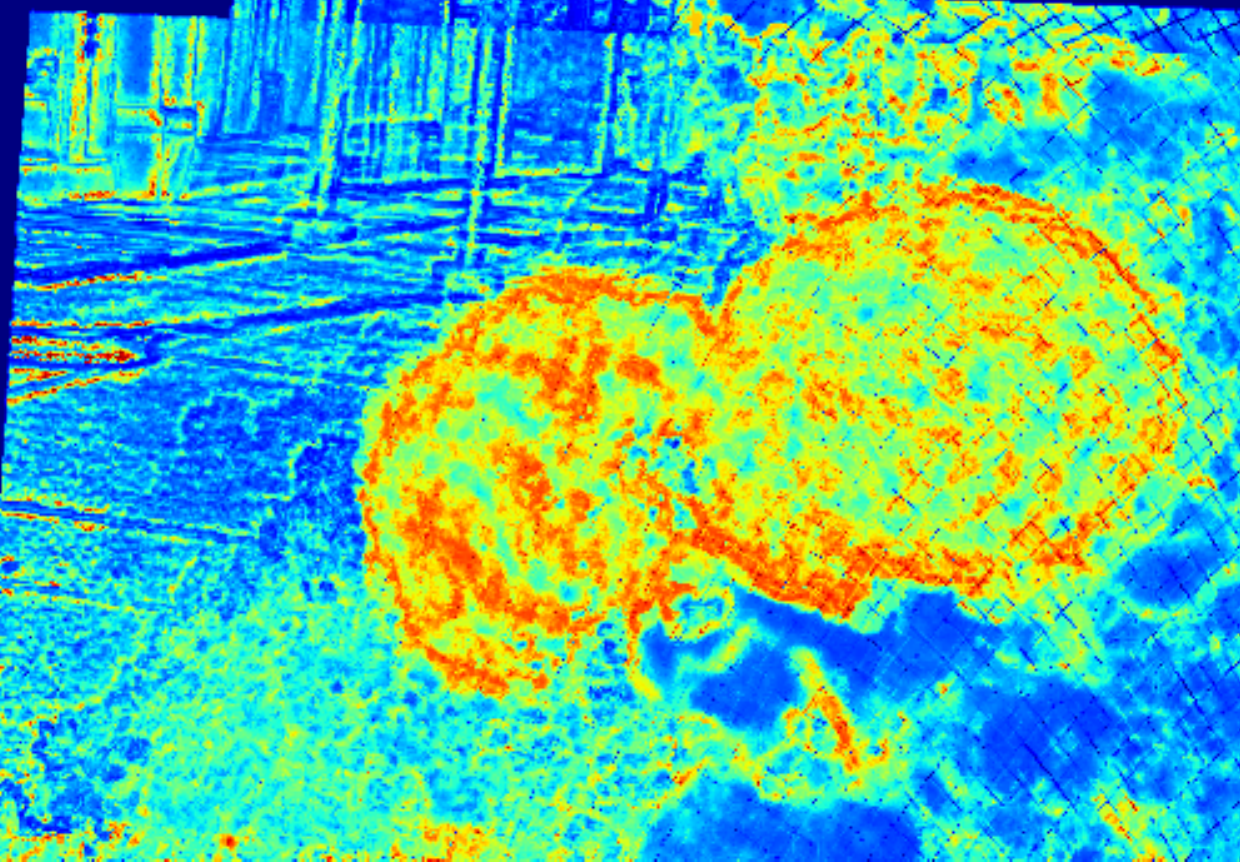}
\end{minipage}
}\end{subfigure}

\vspace{-0.12in}
\begin{subfigure}{
\centering
\begin{minipage}[t]{0.2\textwidth}
\centering
\includegraphics[width=1.3in]{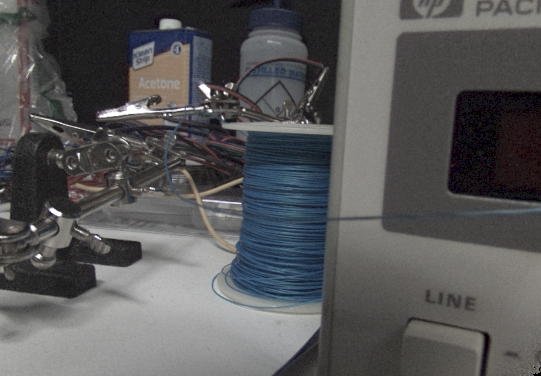}
\end{minipage}

\hspace{-0.2in}
\begin{minipage}[t]{0.2\textwidth}
\centering
\includegraphics[width=1.3in]{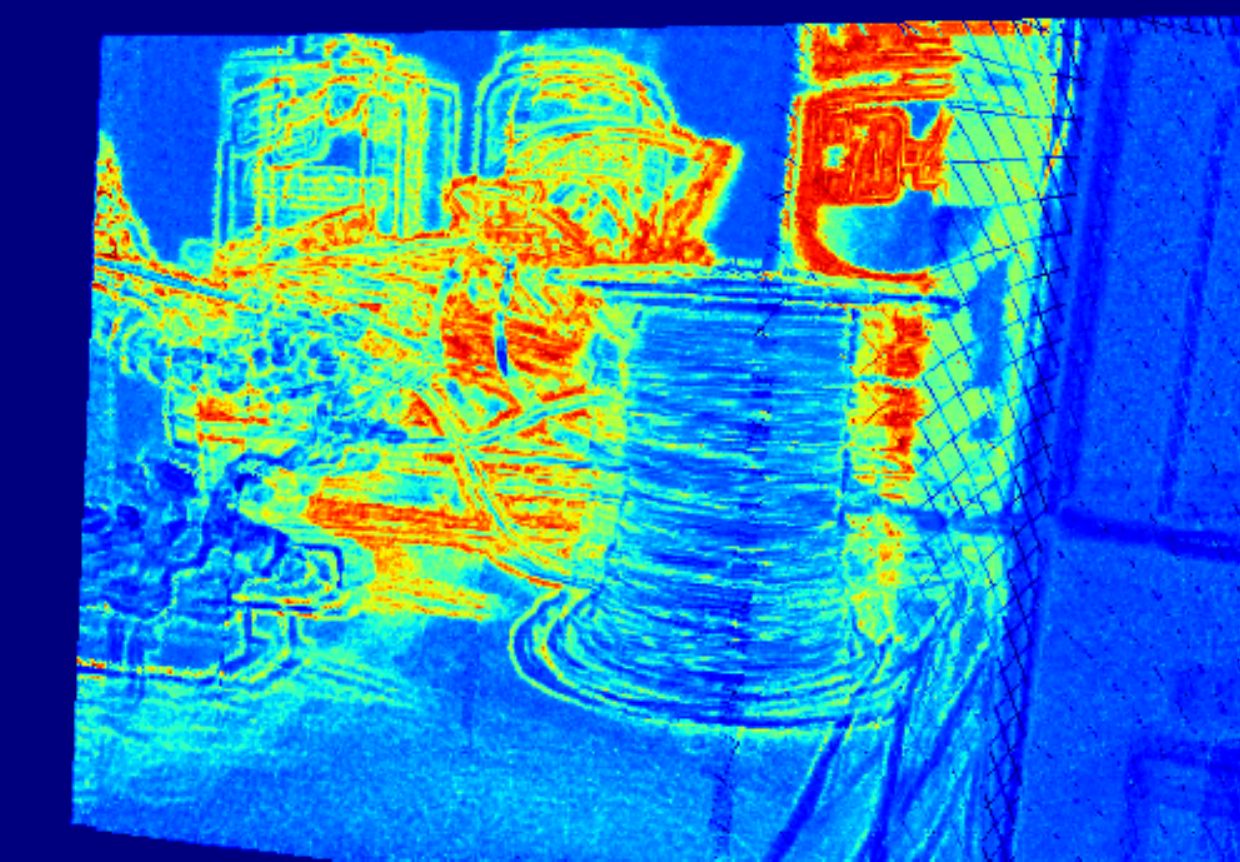}
\end{minipage}

\hspace{-0.2in}
\begin{minipage}[t]{0.2\textwidth}
\centering
\includegraphics[width=1.3in]{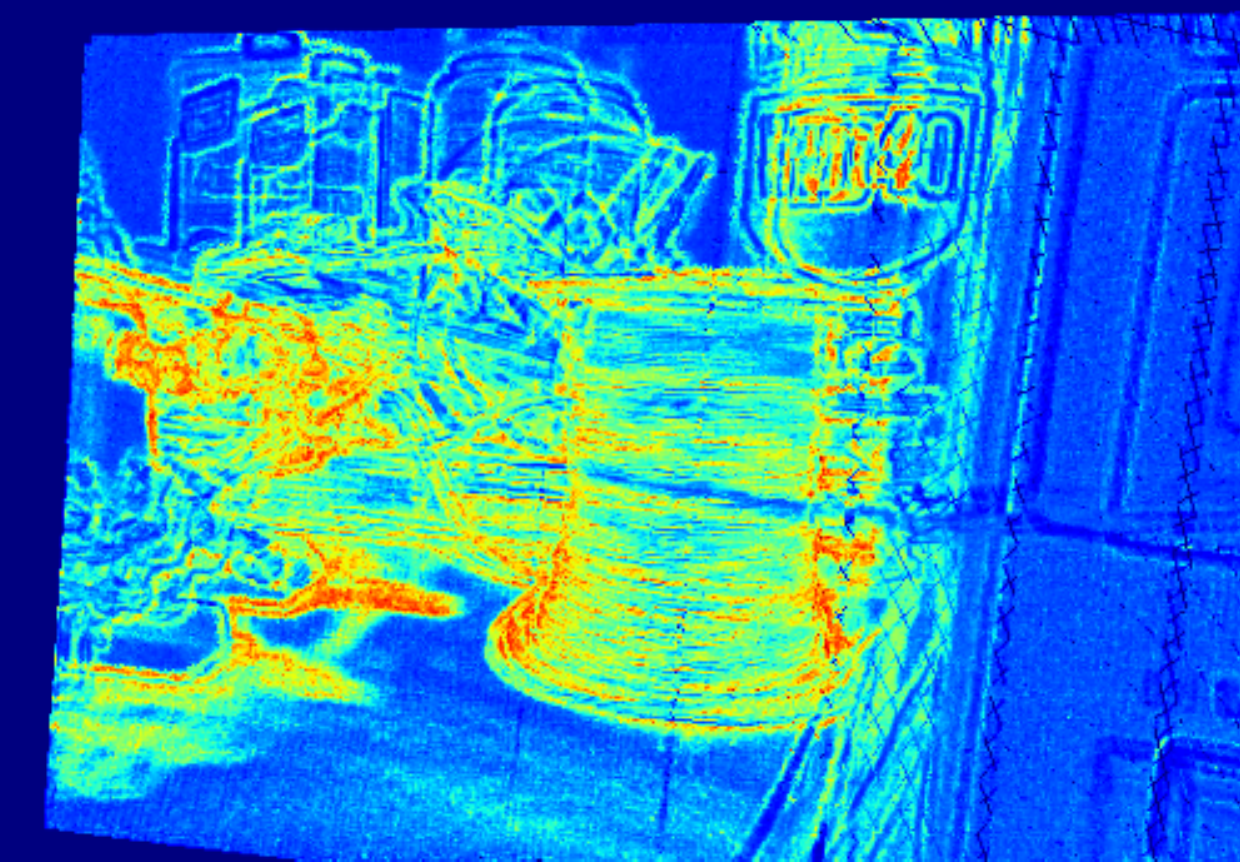}
\end{minipage}

\hspace{-0.2in}
\begin{minipage}[t]{0.2\textwidth}
\centering
\includegraphics[width=1.3in]{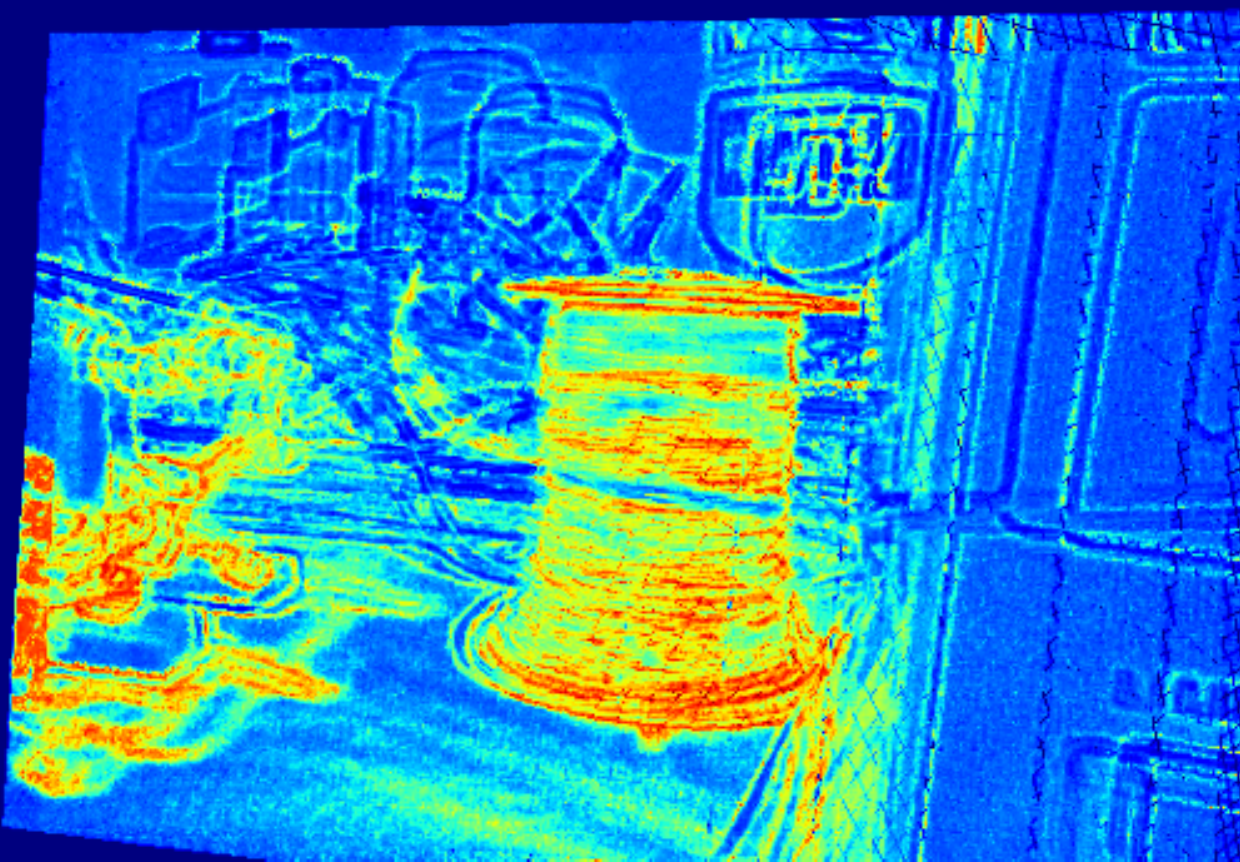}
\end{minipage}

\hspace{-0.2in}
\begin{minipage}[t]{0.2\textwidth}
\centering
\includegraphics[width=1.3in]{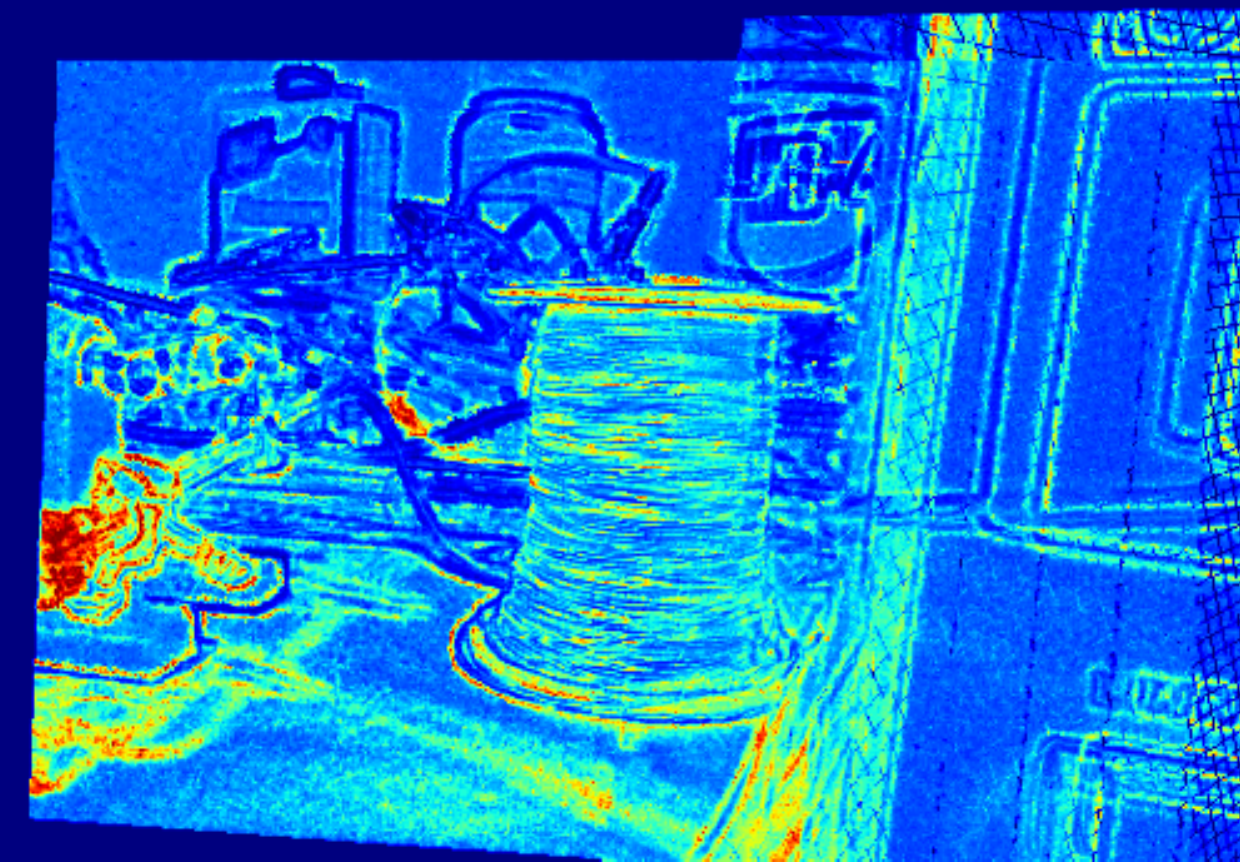}
\end{minipage}
}\end{subfigure}
\caption{Illustration of DPV fusion outcomes. Images on the leftmost column are the target view image $I_{t_0}$. Images starting from the second column to the right show the disparity probability values from disparity planes $d=20$ to 80 (100 planes in total).
Brighter pixels indicate higher probability. }\label{fusionresults}
}\end{figure*}

\subsection{Probability Volume Fusion}\label{Space Sweeping Rays Fusion}
The values in the rescaled probability volume $\mathbf{\bar{U}}_t$ indicate the pixels' probability of having correct disparity at the corresponding plane, \textit{based on the observation from} $\mathbf{L}_t$. As visualized in Fig.~\ref{fig:dispdist}(a), a peak along the disparity axis can be observed for regions with clear textures, which gives strong indication on its correct disparity value. However, for texture-less regions (Fig.~\ref{fig:dispdist}(b)) or near regions with depth discontinues (Fig.~\ref{fig:dispdist}(c)), the probability distributions provide less informative indications.
Nevertheless, when these weak and implicit clues are fused from multiple observation angles and interpreted in a \textit{spatially} and \textit{semantically} congruent manner, we expect a much more confident prediction.
This motivates us to investigate a robust fusion mechanism for accurate target view geometry inference.

We apply \textit{Homography Warping} $\mathcal{H}(d)$ to transform the source volumes $\{\mathbf{\bar{U}}_t\}_{t=1}^{N_\text{src}}$ to the target camera's frustum. The coordinates mapped from the source viewpoint $t$ to the target viewpoint $t_0$ is determined by the planar homography transformation:
\begin{equation}
\mathbf{U}_{t\rightarrow t_0}(x,y,d)=\mathcal{H}(d) \cdot \mathbf{\bar{U}}_t(x,y,d).
\end{equation}
The homography warping operator $\mathcal{H}(d)$ is defined as:
\begin{align}\label{homography}
\mathcal{H}(d)=K_t\cdot R_t\cdot (\mathbf{I} - \frac{(\tau_t-\tau_{t_0} )\cdot {n_{t_0}^T}}{d}  )\cdot R_{t_0}^T \cdot  K_{t_0}^T,
\end{align}
where $n_{t_0}^T$ denotes the principle axis of the target camera frustum, and $\{ {K, R, \tau}\}$, estimated by COLMAP, denote the camera \emph{intrinsics, rotation} and \emph{translation} matrices, respectively. 
As demonstrated in Fig.~\ref{fig:overall}, after the homography warping, all the disparity planes from the volumes $\mathbf{\bar{U}}_t$ are transformed as fronto-parallel planes onto the target view camera's frustum.

To efficiently fuse the source camera's probabilities to the target camera's probability volume, $\mathbf{U}_{t_0}(x,y,d)$ is estimated by fusing the nearest probability bins (as shown in Fig.~\ref{fig:overall}(e)) from $\mathbf{U}_{t\rightarrow t_0}$ via:
\begin{align}\label{raysfusion}
\mathbf{U}_{t_0}(x,y,d) = \frac{\sum_t W_{t}^\text{pos} \cdot W_{t}^\text{dir} \cdot {\mathbf{U}_{t\rightarrow t_0}(x,y,d)}}{N_\text{rays}}.
\end{align}%
Here $N_\text{rays}$ stands for the count of \textit{contributing sources} at the current bin. $W_{t}^\text{pos}$ and $W_{t}^\text{dir}$ are fusion weights that take source camera's location and frustum direction differences into account during fusion: source cameras with larger position and directional differences to the target camera are less reliable and should contribute less during fusion. Both $W_{t}^\text{pos}$ and $W_{t}^\text{dir}$ are normalized by the softmax function with euclidean distances from all source cameras. Fig.~\ref{fusionresults} gives visual demonstrations of the fused 3D data volume $\mathbf{U}_{t_0}$. 

\textbf{Remark.} Compared with MVSNet\cite{yao2018mvsnet}, or the other multi-view stereo methods that also perform homography warping on images or features, we directly work on probability distributions. Our method has two advantages: first, the generation of source view features requires larger memory and intensive computation, which usually require down-sampling of feature resolutions and thus limit the quality of estimated geometry; second, fusion of probability distribution is much more robust compared to direct fusion of image features. It combines weak probability clues into stronger ones via the weighted fusion mechanism as in Eq. (\ref{raysfusion}). This enables our framework to deal with larger baseline scenarios. 

\subsection{Frustum Voxel Filtering and Attention-Guided Multi-scale Residual Fusion.} \label{fvprocessing}
The fused 3D data volume $\mathbf{U}_{t_0}(x,y,d) \in \mathbb{R}^{H_0\times W_0 \times N_\text{dp}}$ sub-divides the camera's visible space into frustum voxels (FV).
As can be observed from Fig.~\ref{fusionresults}, there are explicit peaks and less \textit{smearing effect}\footnote{Each FV stores the disparity probability value that is homography warped and fused from multi-LF observations. The smearing effect is caused by the dispersion of probability peaks during warping.} over regions with edges and textures.
However, a significant amount of FVs are noisy and without explicit values over texture-less regions. In order to propagate correct probability values to the relevant FVs, we utilize a 3D-UNet structure (network detail specified in TABLE~\ref{3dunet}) to further filter the frustum data.
The output will be a deeply regularized volume $\mathbf{\hat{U}}_{t_0}$ as shown in Fig.~\ref{fig:overall}(f), over which, a preliminary disparity map $D_{t_0}$ will be generated as weighted sum over all hypothesized planes:

\begin{align}\label{rawdisparitymap}
D_{t_0}(x,y)=\sum_{d} \mathbf{\hat{U}}_{t_0}(x,y,d)\times d.
\end{align}\vspace{-0.4cm}

To further improve the quality of $D_{t_0}$ by ensuring boundary alignment and semantic consistency with the target view RGB capture $I_{t_0}$,
an Attention-Guided Multi-scale Residual Fusion Module is designed which progressively combines encoded features from the target image $F_{t_0}^i$ with the features from the preliminary disparity map $F_\text{disp}^i$ in a multi-scale and coarse-to-fine manner.
The structure of the module is shown in Fig. \ref{fig:overall}(c).  \textit{ResNeXt-101-WSL} \cite{wslimageseccv2018, Ranftl2020Towards} has been adopted as feature encoder.
As specified from Eq.~(\ref{residual_scale})~to~(\ref{fusion}), the final refined disparity map $\hat{D}_{t_0}$ at the target view is generated by fusing multi-scale residuals $R^{i}_{t_0} $, $R^{i}_\text{disp}$ (upsampled by $\Gamma_\text{up}$) weighted based on the predicted attention masks $W^{i}_{t_0}$ and $W^{i}_\text{disp}$, which help to locate informative regions for efficient residual fusion.\vspace{-0.4cm}

\begin{align}
\label{residual_scale}
R^{i}_\text{disp} &= \text{conv}(\text{ReLU}(\text{conv}(\text{ReLU}(F^{i}_\text{disp})))) + F^{i}_\text{disp},\\[0.2cm]
\label{residual_color}
R^{i}_\text{rgb} &= \text{conv}(\text{ReLU}(\text{conv}(\text{ReLU}(F^{i}_\text{rgb})))) + F^{i}_\text{rgb},\\[0.2cm]
\label{atten_scale}
W^{i}_\text{disp} &= \text{ReLU(conv(ReLU(}R^{i}_\text{disp}))),\\[0.2cm]
\label{atten_color}
W^{i}_\text{rgb} &= \text{ReLU}(\text{conv}(\text{ReLU}(R^{i}_\text{rgb}))),\\[0.2cm]
\label{upsample}
\Gamma_\text{up}[\cdot] &=\text{Interpolate}(\text{conv}(\cdot)), \\[0.2cm]
\label{fusion}
\hat{D}^{i}_{t_0} &= W^{i}_\text{disp} \odot R^{i}_\text{disp} +W^{i}_\text{rgb}\odot R^{i}_\text{rgb} + \Gamma_\text{up}[\hat{D}^{i-1}_{t_0}],
\end{align}
where $i$ is the scale index of the decoder. The operator $\Gamma_\text{up}$ upsamples the feature by 2 using a bilinear interpolator.

\begin{table}[t]
\centering

\caption{Network Structure of the Frustum Voxel Filtering Module. Here $k$ denotes 3D kernel size (as $k\times k\times k$), $s$ denotes stride, $d$ denotes number of padding, $C_\text{in}$ denotes the number of input feature channel, $C_\text{out}$ denotes the number of output feature channel.}
\label{3dunet}
\begin{tabular}{lllllllll}

\hline
Layer   & $k$ & $s$ & $d$  & $C_\text{in}$ & $C_\text{out}$ & Input & Output\\ \rowcolor{mygray}
\hline
 Conv 0\_1 & 3 & 1 & 1   &1 &10 & $\mathbf{U}_{t_0}$ & $V_\text{0\_1}$\\
 Conv 1\_0 & 3 & 2 & 2   &10 &20 & $V_\text{0\_1}$ & $V_\text{1\_0}$\\ \rowcolor{mygray}
 Conv 2\_0 & 3 & 2 & 2  &20 &40 & $V_\text{1\_0}$ & $V_\text{2\_0}$\\
 Conv 3\_0 & 3 & 2 & 1  &40 &120 & $V_\text{2\_0}$ & $V_\text{3\_0}$\\ \rowcolor{mygray}
 Conv 1\_1 & 3 & 1 & 1  &20 &20 & $V_\text{1\_0}$ & $V_\text{1\_1}$\\
 Conv 2\_1 & 3 & 1 & 1  &40 &40 & $V_\text{2\_0}$ & $V_\text{2\_1}$ \\  \rowcolor{mygray}
 Conv 3\_1 & 3 & 1 & 1  &120 &120 &$V_\text{3\_0}$ & $V_\text{3\_1}$ \\
 Conv 3\_2 & 3 & 2 & 1  &120 &40 & $V_\text{3\_1}$ & $V_\text{3\_2}$\\  \rowcolor{mygray}
 Conv 2\_2 & 3 & 2 & 2  &40 &20 & $V_\text{2\_1} + V_\text{3\_2}$ & $V_\text{2\_2}$\\
 Conv 1\_2 & 3 & 2 & 2  &20 &10 & $V_\text{2\_2} + V_\text{1\_1}$ & $V_\text{1\_2}$\\  \rowcolor{mygray}
 Conv 0\_2 & 1 & 3 & 1  &10 &1 & $V_\text{0\_1} + V_\text{1\_2}$ & $\mathbf{\hat{U}}_{t_0}$ \\
\hline
\end{tabular}
\end{table}

\subsection{Disparity Field Synthesis and Immersive Rendering}\label{lightfieldsynthesissection}

The final step of our framework is to utilize the refined disparity map to raise the target view image $I_{t_0} \in \mathbb{R}^{H_0\times W_0\times C_0}$ into a locally immersive LF: $\mathbf{\hat{L}}_{t_0}\in\mathbb{R}^{H_0\times W_0\times C_0\times M_0 \times N_0}$.
This involves two sub-steps: disparity field synthesis and immersive content rendering.

\textbf{\textit{Disparity Field Synthesis}}.
The goal of this module is to raise the disparity map $\hat{D}_{t_0} \in \mathbb{R}^{H_0\times W_0}$ into a densely-sampled immersive field $\mathbf{F}\in \mathbb{R}^{H_0\times W_0 \times M_0 \times N_0}$. Each angular slice of $\mathbf{F}$ represents the parallax of the sub-aperture view with respect to $I_{t_0}$.
As shown in Fig.~\ref{fig:overall}(d), the disparity field synthesis network takes $\hat{D}_{t_0}\in \mathbb{R}^{H_0\times W_0\times 1}$ as input, and first uses a 2D convolution layer to upsample the channel dimension of $\hat{D}_{t_0}$ from 1 to $M_0\times N_0$. The convolution layer learns the spatial-angular parallax correlations among the upsampled angular channel slices, and they are further regularized by a Spatial-Angular Alternating (SAA) Convolution Module \cite{8561240, article, yeung2018fast}, which finally generates the spatial-angular consistent disparity field.
The SAA Convolution Module consists of a set of Spatial-Angular Convolution Blocks; each block performs pseudo-4D convolutions by alternatively carrying out 2D convolutions on the spatial and angular dimensions of the 4D data. The SAA module deeply regularizes the implicit structures of the disparity field in a computationally efficient manner.

\textbf{\textit{Immersive Content Rendering}}. Compared with previous methods of synthesizing LF using a sparse set of input views \cite{LearningViewSynthesis, srinivasan2017learning}, synthesizing an immersive LF from a single image is much more challenging. Nonetheless, we can fully exploit the accurate scene geometric prior $\mathbf{F}$ to backward warp pixels from the target view $I_{t_0}$ to other SAIs as illustrated in Eq. (\ref{eq:warping}).
\begin{equation}\label{eq:warping}
\mathbf{L}_{t_0}(\mathbf{x}, \mathbf{v}) = I_{t_0}(\mathbf{x} + \mathbf{F}(\mathbf{x}, \mathbf{v}) \times \mathbf{v}),
\end{equation}
where $\mathbf{x}$ is the 2D coordinates in $I_{t_0}$, $\mathbf{v} \in [0, M_0\times N_0]$ is the angular index of the SAI, $\mathbf{v}$ is the angular offsets between the $\mathbf{v}$-th SAI with respect to the central view, $\mathbf{L}_{t_0}$ is the synthesized preliminary LF. The preliminary LF is further refined by adding the residual from a second SAA module with the same network structure as the one in the disparity field network. Consequently, the output of the Immersive Content Rendering is a refined LF $\mathbf{\hat{L}}_{t_0}$.

\textbf{Remark.} We have employed the methodology of first synthesizing the disparity field and based on which, subsequently synthesizing the light field. SAA module has been applied twice, which efficiently \textit{\ul{re-regularizes}} the structure of the high dimensional LF data. The high-quality disparity estimation $\hat{D}_{t_0}$ is also the main reason for high-quality LF synthesis.

\begin{figure*}[t]{
\centering

\begin{subfigure}{
\centering
\hspace{+0.01in}
\begin{minipage}[t]{0.2\textwidth}
\centering
\includegraphics[width=1.3in]{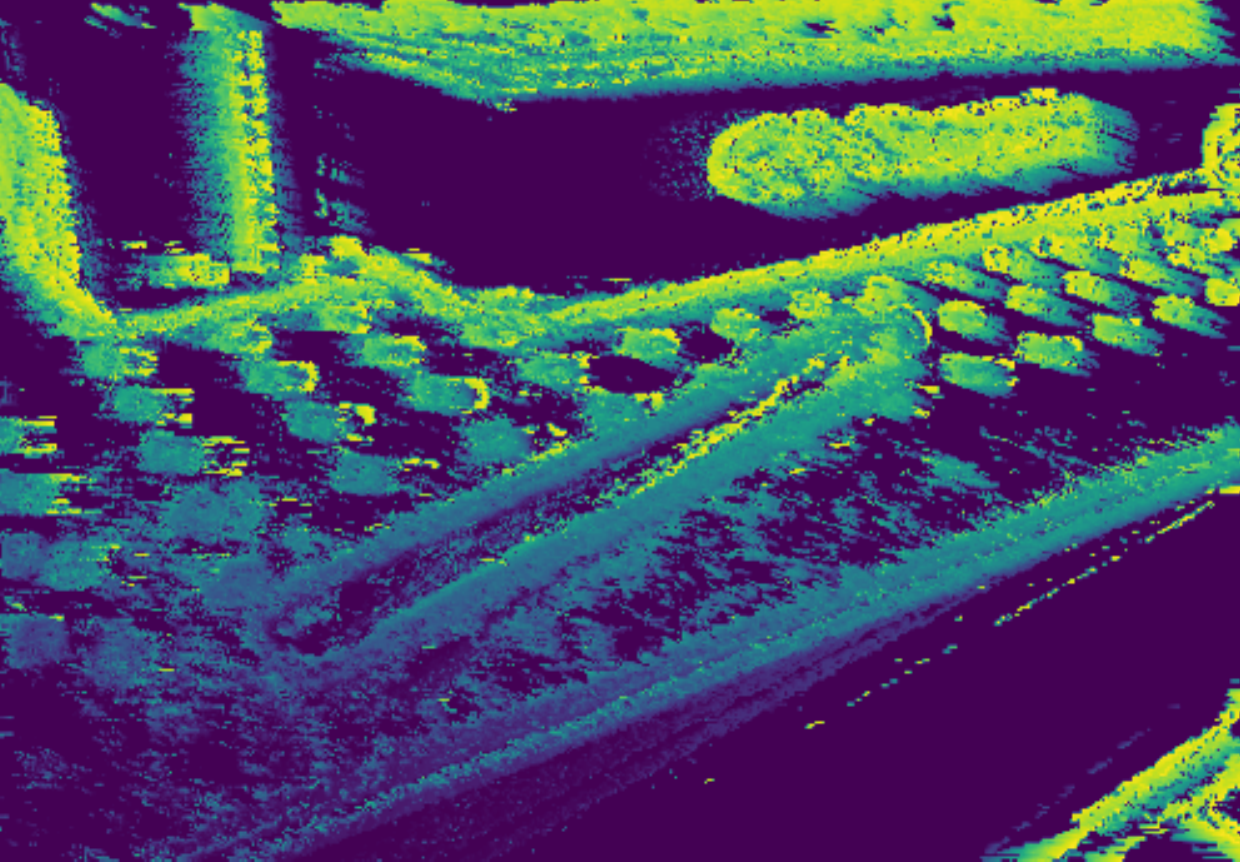}
\end{minipage}
\hspace{-0.2in}
\begin{minipage}[t]{0.2\textwidth}
\centering
\includegraphics[width=1.3in]{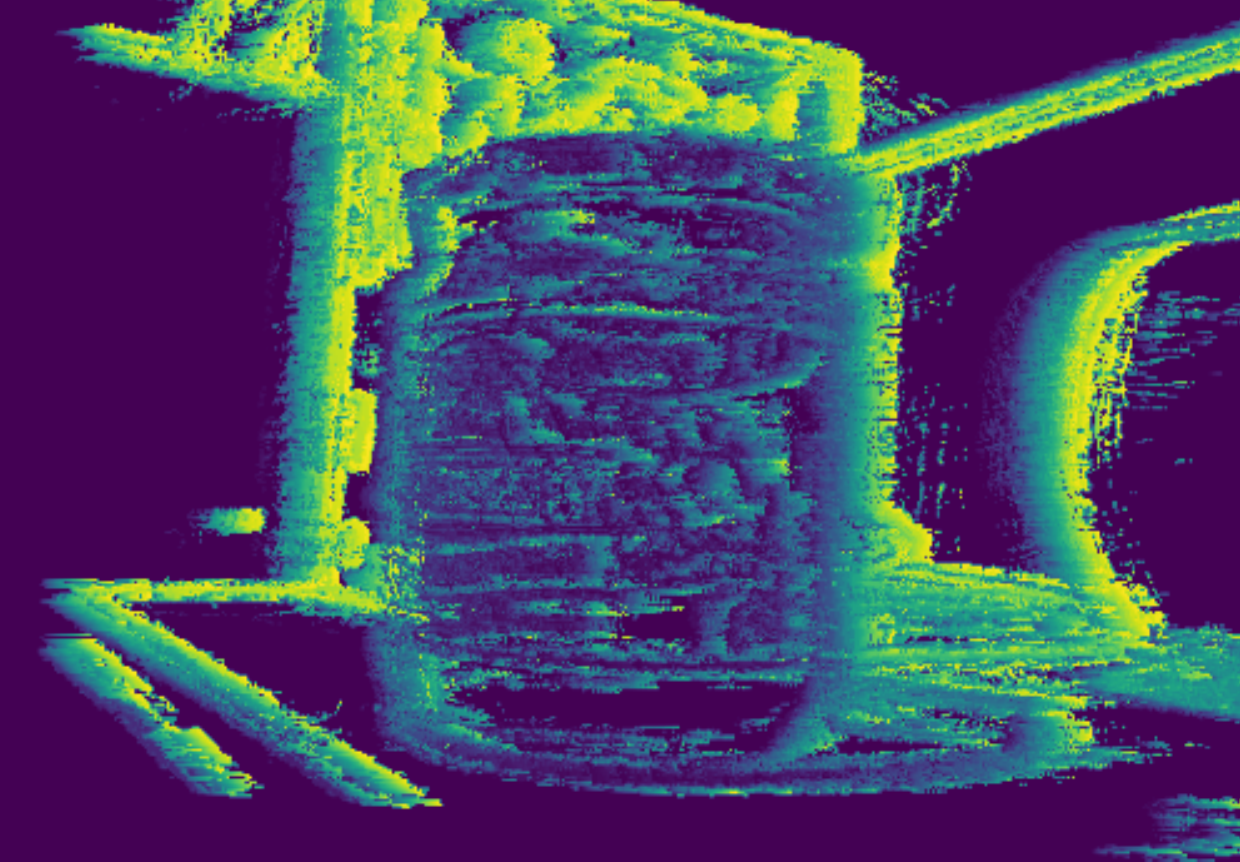}
\end{minipage}
\hspace{-0.2in}
\begin{minipage}[t]{0.2\textwidth}
\centering
\includegraphics[width=1.3in]{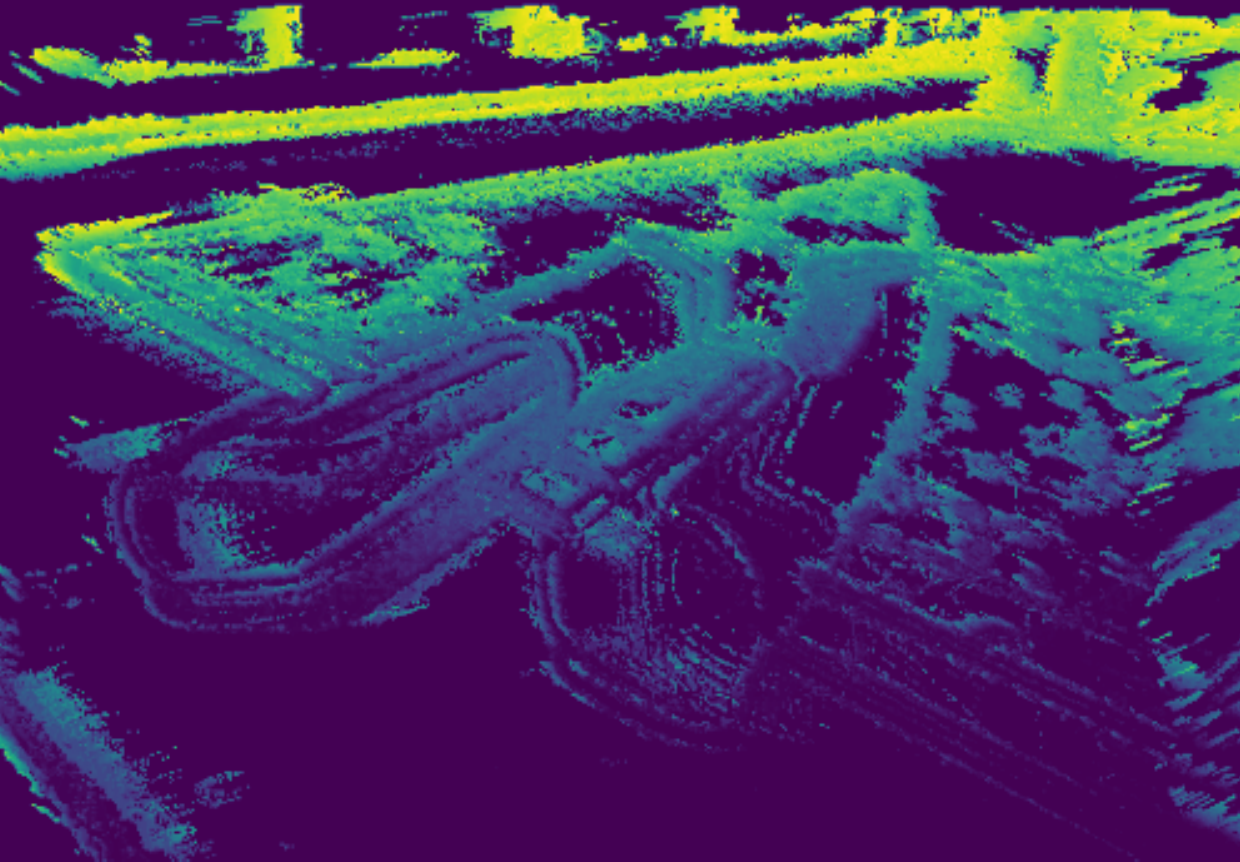}
\end{minipage}
\hspace{-0.2in}
\begin{minipage}[t]{0.2\textwidth}
\centering
\includegraphics[width=1.3in]{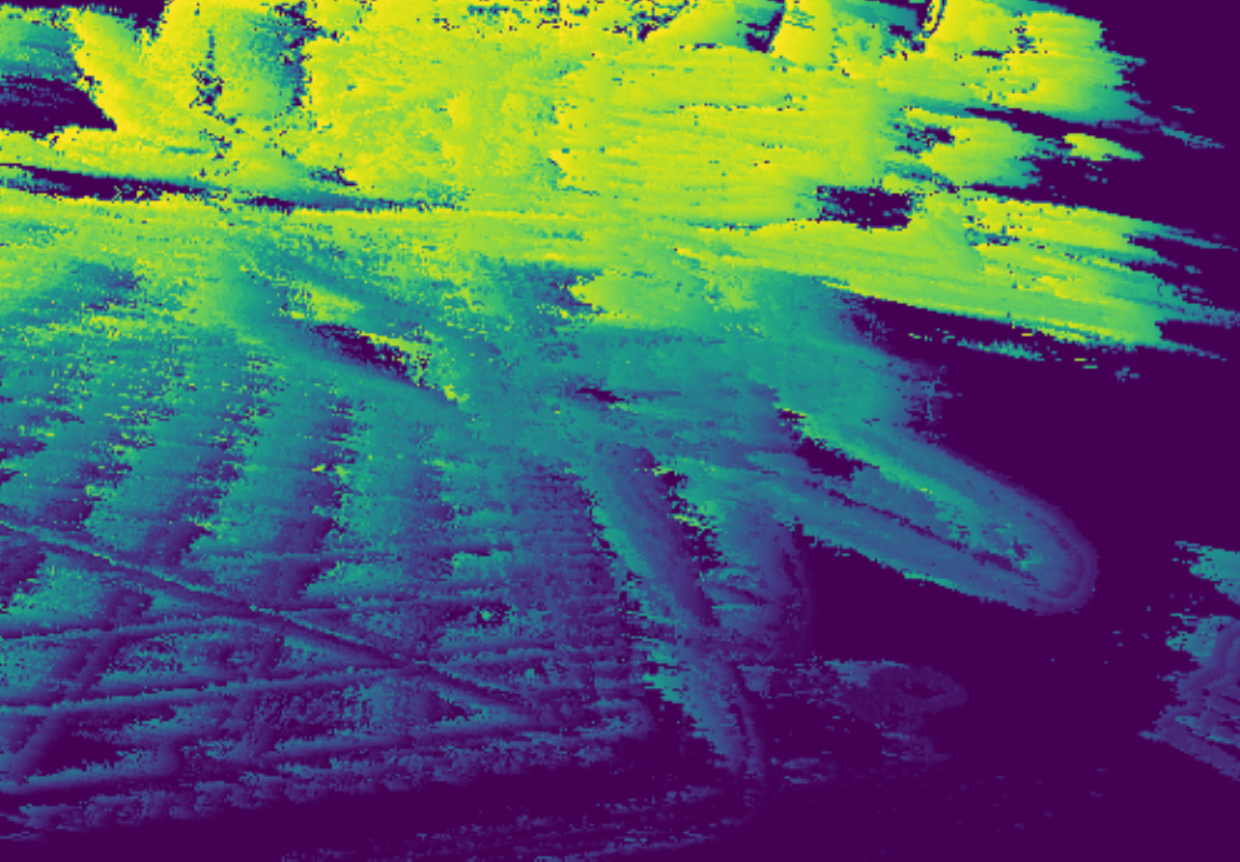}
\end{minipage}
\hspace{-0.2in}
\begin{minipage}[t]{0.2\textwidth}
\centering
\includegraphics[width=1.3in]{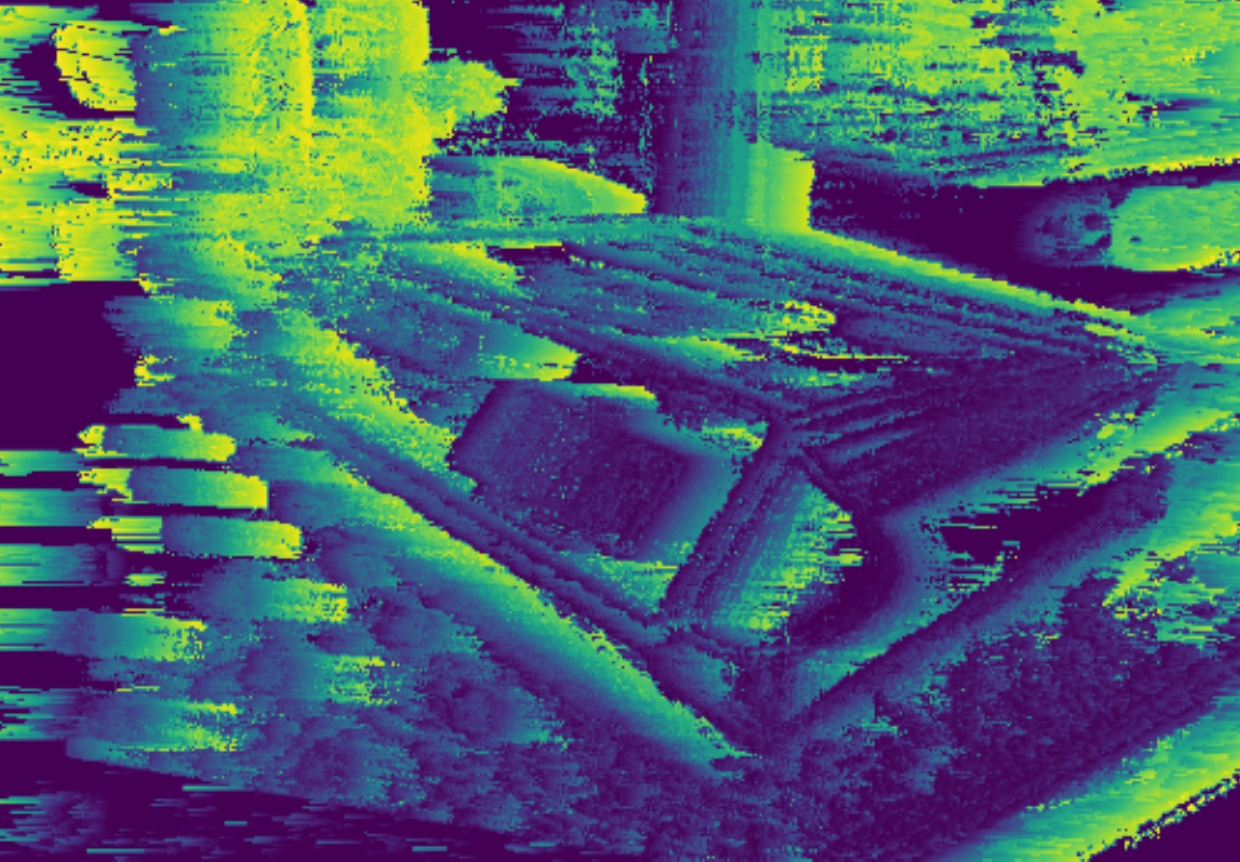}
\end{minipage}
}\end{subfigure}
\vspace{-0.3cm}

\begin{subfigure}{
\centering
\begin{minipage}[t]{0.2\textwidth}
\centering
\includegraphics[width=1.3in]{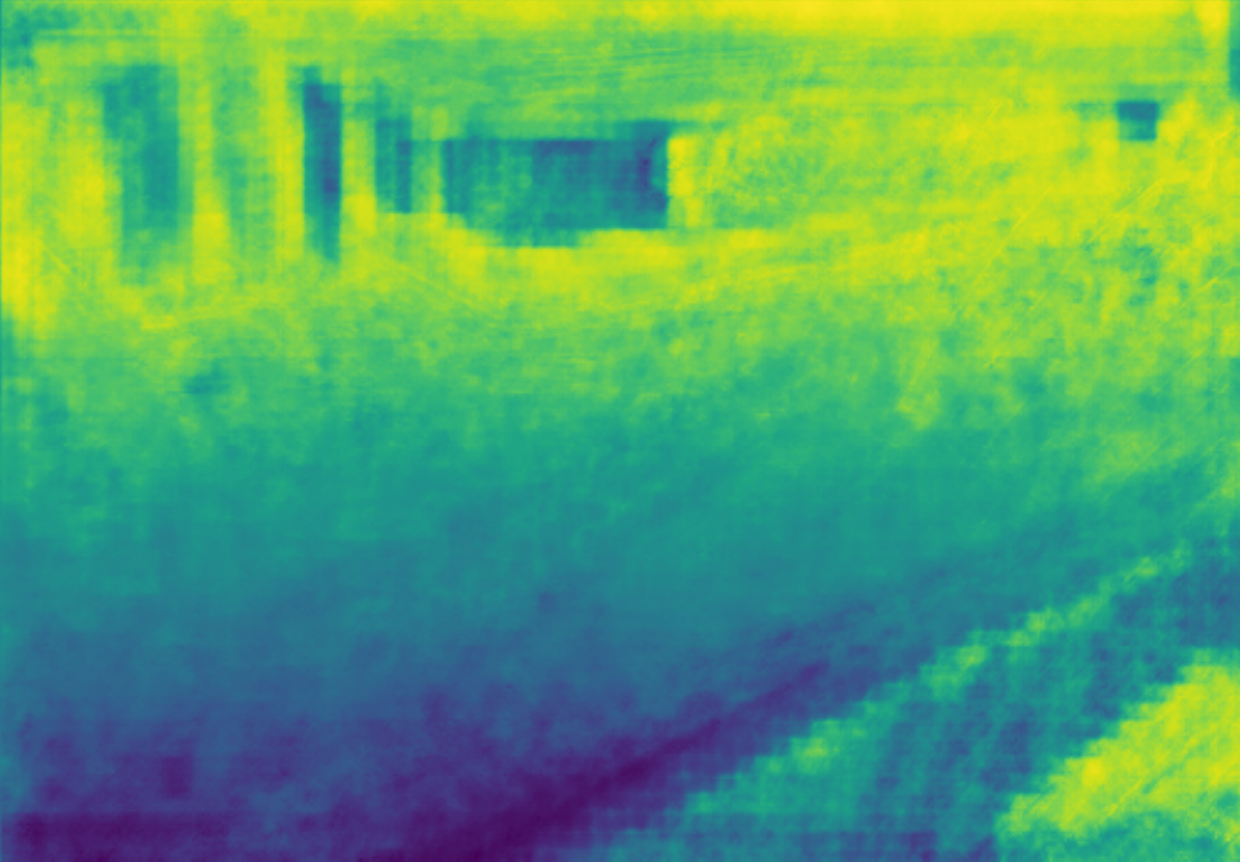}
\end{minipage}
\hspace{-0.2in}
\begin{minipage}[t]{0.2\textwidth}
\centering
\includegraphics[width=1.3in]{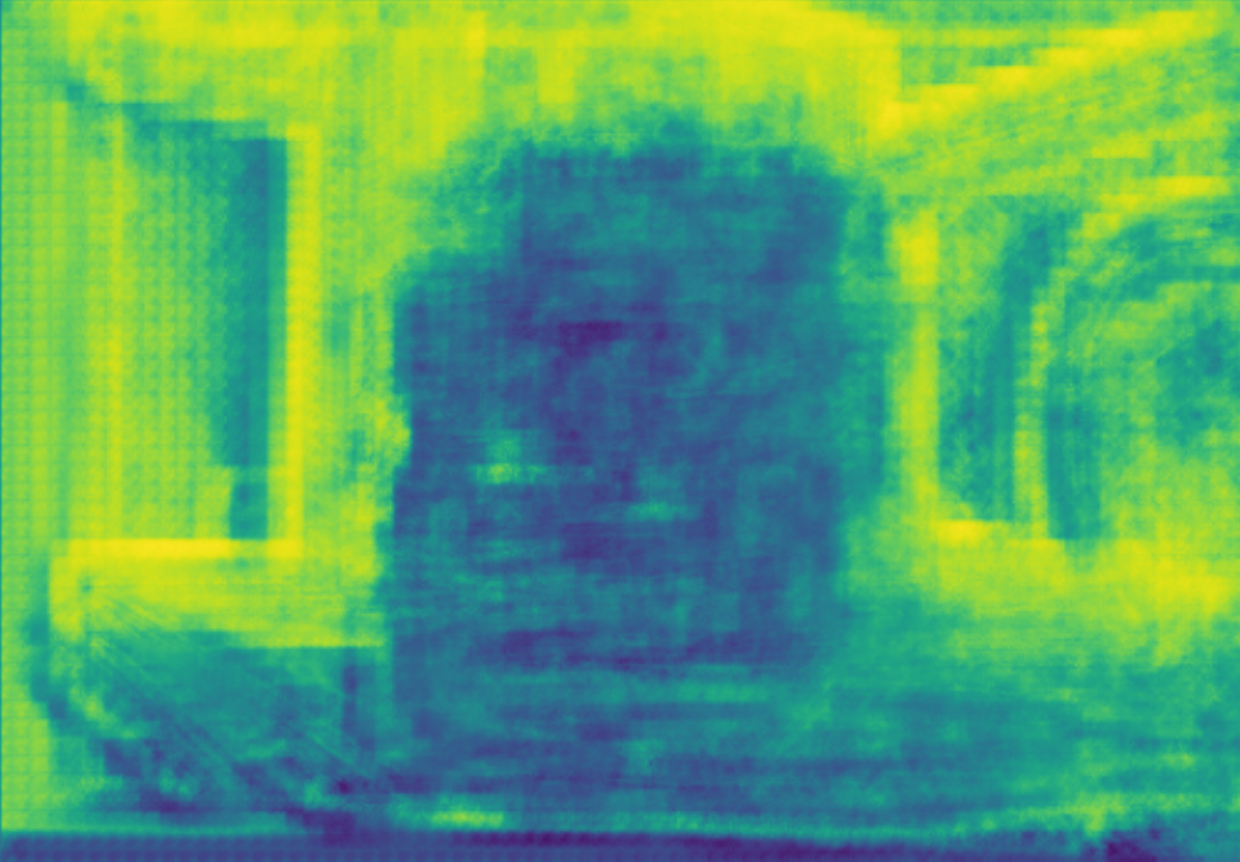}
\end{minipage}
\hspace{-0.2in}
\begin{minipage}[t]{0.2\textwidth}
\centering
\includegraphics[width=1.3in]{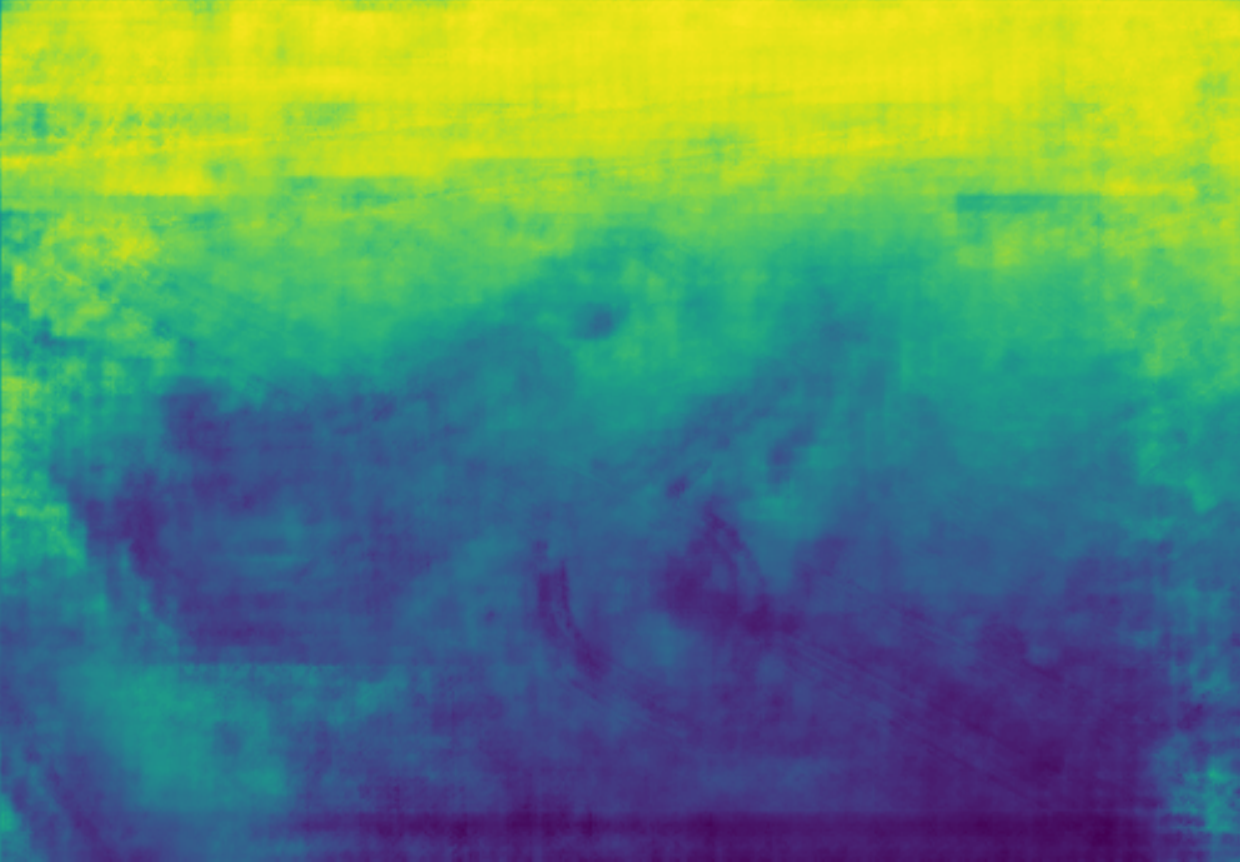}
\end{minipage}
\hspace{-0.2in}
\begin{minipage}[t]{0.2\textwidth}
\centering
\includegraphics[width=1.3in]{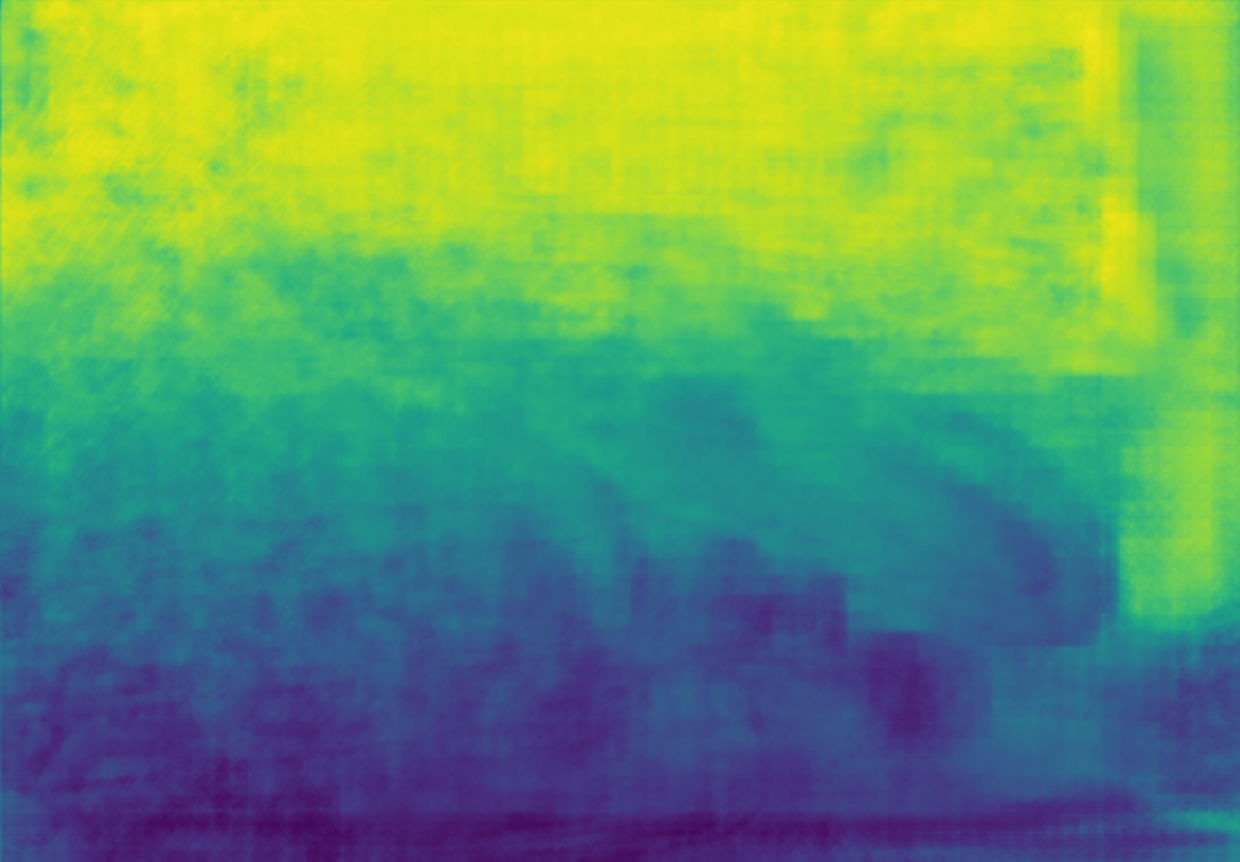}
\end{minipage}
\hspace{-0.2in}
\begin{minipage}[t]{0.2\textwidth}
\centering
\includegraphics[width=1.3in]{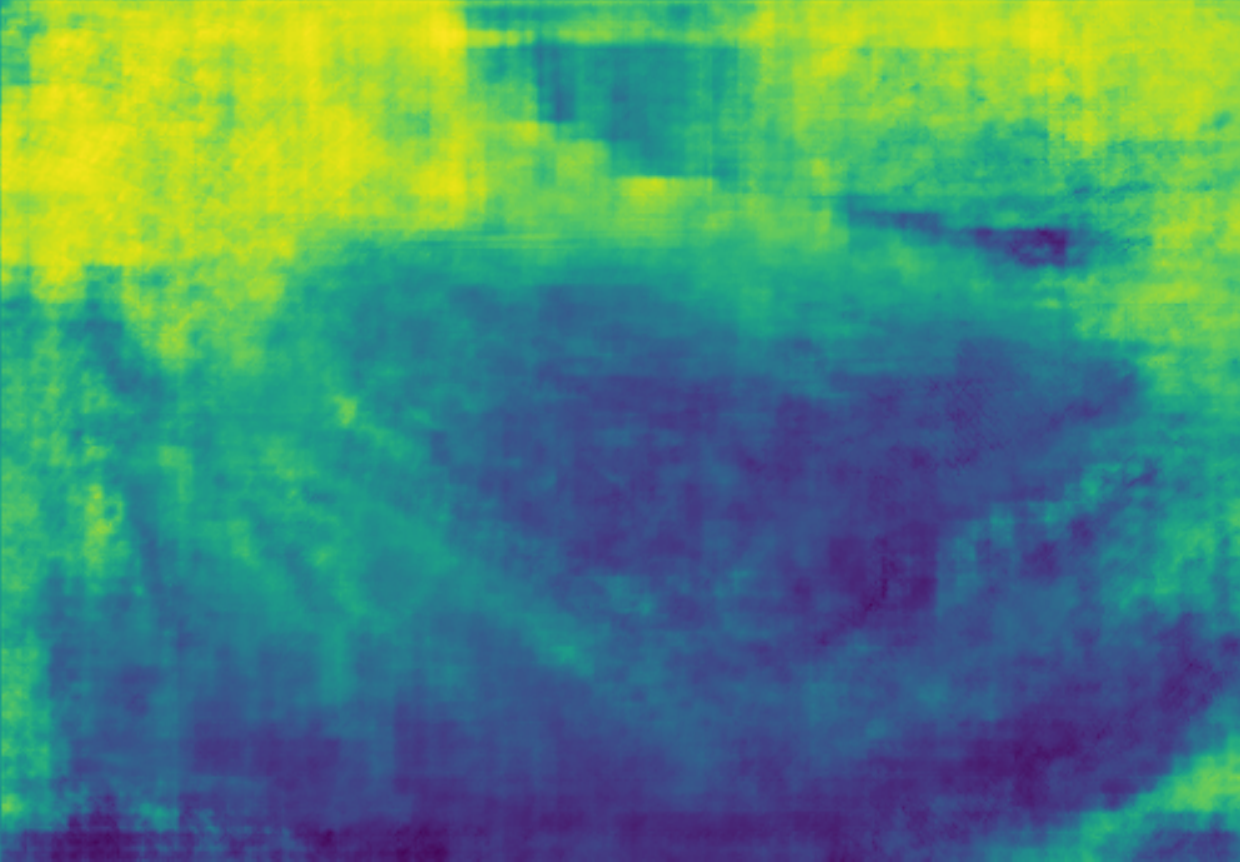}
\end{minipage}
}\end{subfigure}

\vspace{-0.3cm}
\begin{subfigure}{
\centering
\begin{minipage}[t]{0.2\textwidth}
\centering
\includegraphics[width=1.3in]{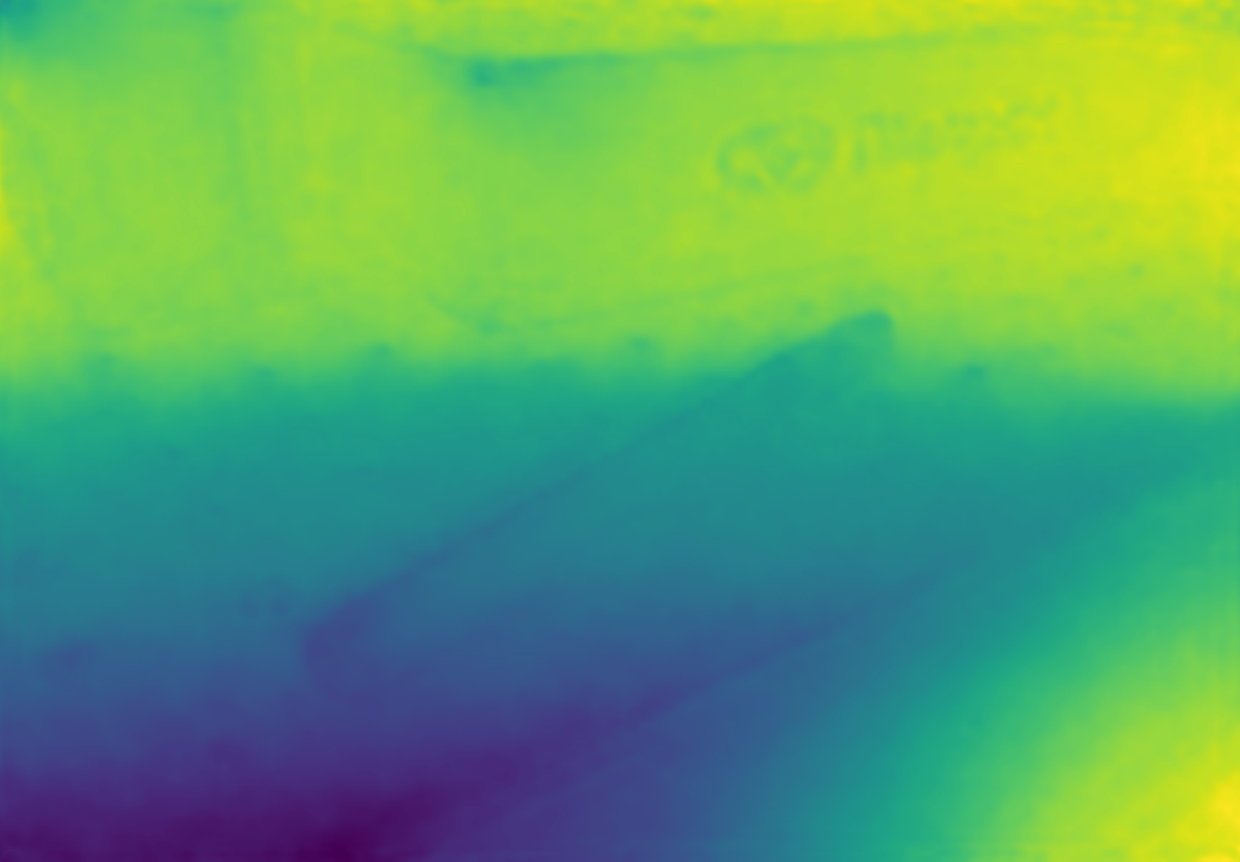}
\end{minipage}
\hspace{-0.2in}
\begin{minipage}[t]{0.2\textwidth}
\centering
\includegraphics[width=1.3in]{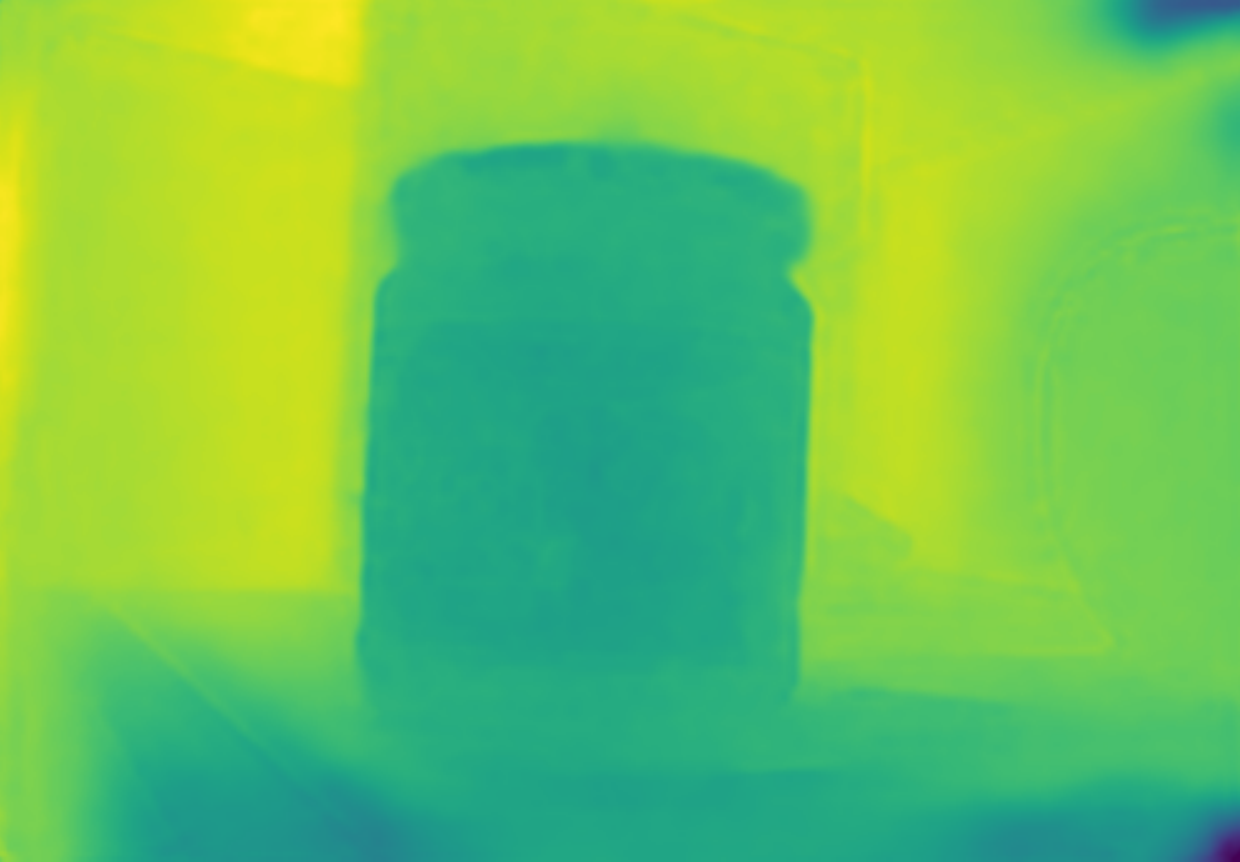}
\end{minipage}
\hspace{-0.2in}
\begin{minipage}[t]{0.2\textwidth}
\centering
\includegraphics[width=1.3in]{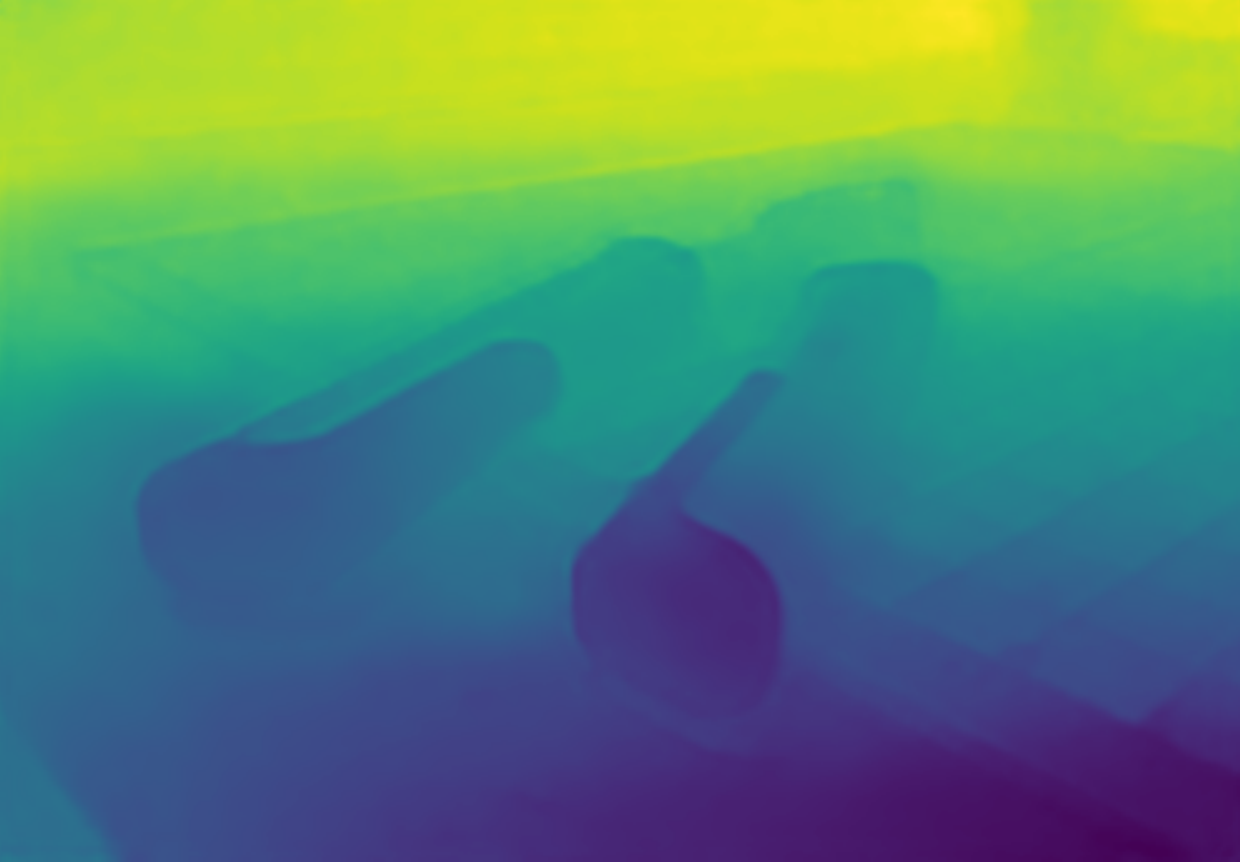}
\end{minipage}
\hspace{-0.2in}
\begin{minipage}[t]{0.2\textwidth}
\centering
\includegraphics[width=1.3in]{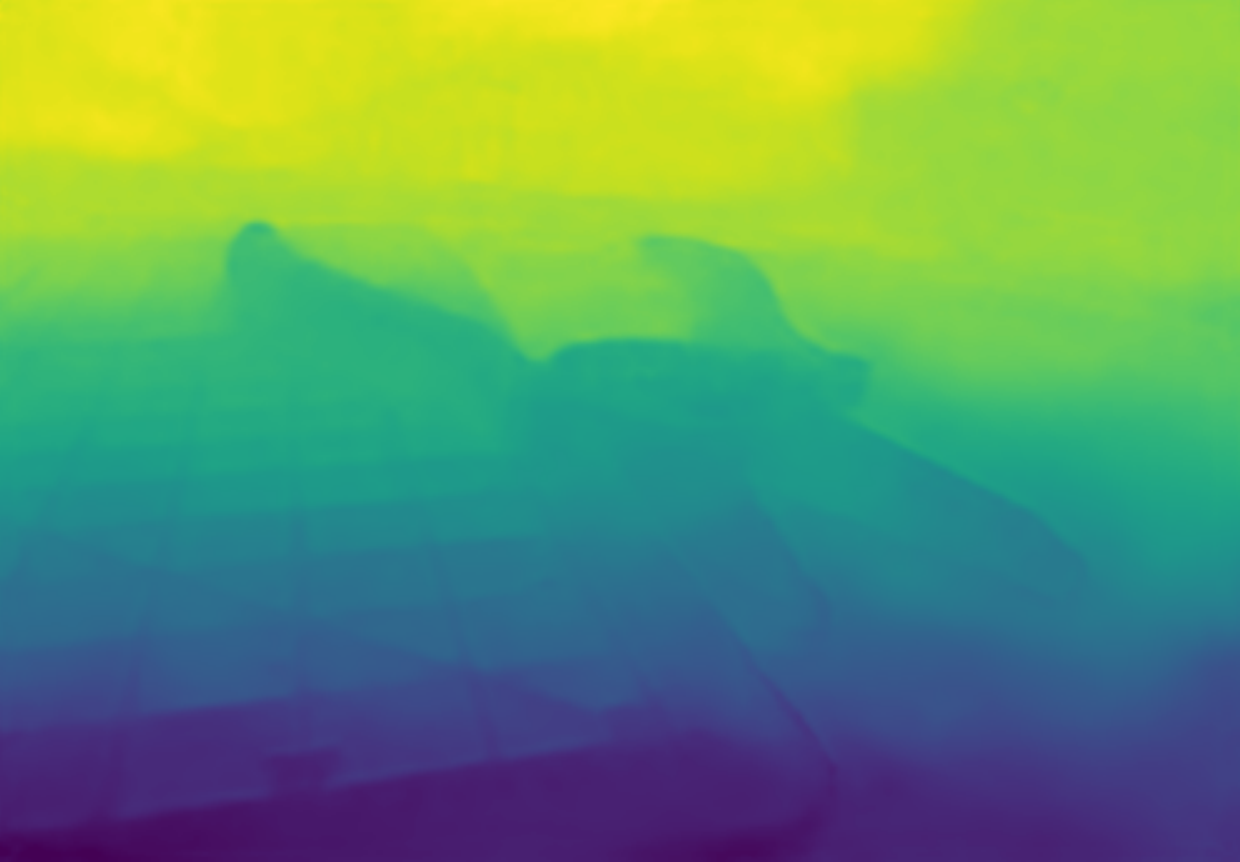}
\end{minipage}
\hspace{-0.2in}
\begin{minipage}[t]{0.2\textwidth}
\centering
\includegraphics[width=1.3in]{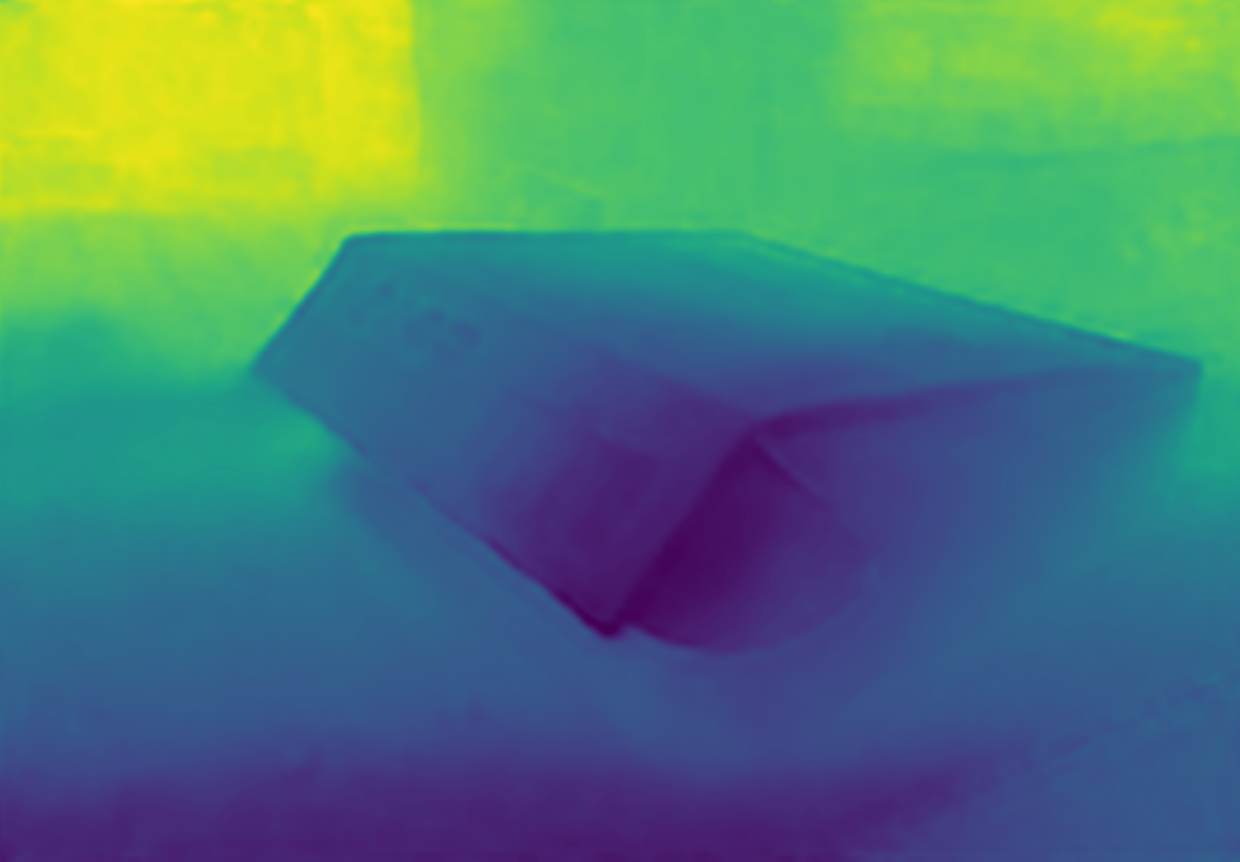}
\end{minipage}
}\end{subfigure}
\caption{The images on the first row are disparity maps directly inferred from $\mathbf{\bar{U}_t}$ before being processed by the Frustum Voxel Filtering Module. We can observe lots of blank space (in dark blue) without disparity values in the first row. The images on the second row are the preliminary disparity maps $D_{t_0}$ inferred from the regularized DPV $\mathbf{\hat{U}}_{t_0}$, which are much better in completeness. The images on the third row are the final disparity estimations $\hat{D}_{t_0}$ which combines texture and semantic constraints from the target view image $I_{t_0}$.}
\label{ablation_study}\vspace{-0.4cm}
}\end{figure*}

\subsection{Model Implementation Details}

\textit{\textbf{The Loss Function:}}
The training of the FV Filtering Module and the Attention-Guided Multi-scale Residual Fusion Module are supervised by $\mathcal{L}_\text{fv}$ and $\mathcal{L}_\text{fused}$, both of which are calculated as the sum of multi-scale Mean Square Error (MSE) between $D_{t_0}$ and $\hat{D}_{t_0}$ with respect to the ground truth disparity $D_{g}$, as shown in Eq. (\ref{unet_loss}) and (\ref{final_loss}).
The final loss for disparity estimation $\mathcal{L}_\text{final}$ is the sum of $\mathcal{L}_\text{fv}$ and $\mathcal{L}_\text{fused}$ as shown in Eq. (\ref{summed_loss}).
\begin{align}\label{unet_loss}
 \mathcal{L}_\text{fv}(D_{t_0},D_{g}) = \sum_{i=1}^{N_\text{scale}} \text{MSE}(D_{t_0}^{i},D_{g}^{i}),
\end{align}\vspace{-0.2cm}
\begin{align}\label{final_loss}
\mathcal{L}_\text{fused}(\hat{D}_{t_0},D_{g}) = \sum_{i=1}^{N_\text{scale}} \text{MSE}(\hat{D}_{t_0}^{i},D_{g}^{i}),
\end{align}\vspace{-0.2cm}
\begin{align}\label{summed_loss}
\mathcal{L}_\text{final} = \mathcal{L}_\text{fv}(D_{t_0}, D_{g}) + \lambda_\text{fv}\mathcal{L}_\text{fused}(\hat{D}_{t_0},D_{g}).
\end{align}
Here $i$ is the scale index for the residual feature fusion module (with $N_\text{scale}$ scales in total); $D_g$ is the ground truth disparity; and $\lambda_\text{fv}$ is the balance weight for the loss $\mathcal{L}_\text{fused}$.

The Light Field Synthesis Module is supervised by the composite loss $\mathcal{L}_\text{lf}$ of MSE, Mean Absolute Error (MAE) and the epipolar-plane image (EPI) losses between the synthesized LF $\mathbf{\hat{L}}_{t_0}$ and the ground truth LF $\mathbf{L}_\text{gt}$:
\begin{equation}\label{lf_summed_loss}
\mathcal{L}_\text{lf} = \text{MSE}(\mathbf{L}_\text{gt},\mathbf{\hat{L}}_{t_0}) + \lambda_1 \text{MAE}(\mathbf{L}_\text{gt},\mathbf{\hat{L}}_{t_0})+ \lambda_2 \text{EPI}(\mathbf{L}_\text{gt},\mathbf{\hat{L}}_{t_0}).
\end{equation}
Here $\lambda_1, \lambda_2$ are the balance weights for respective losses.

\textit{\textbf{The Dataset Details:}}
For model training and evaluation, we used the \textit{Stanford Lytro Multi-view Light Field Dataset} (MVLF) \cite{dansereau2019liff}, which contains a set of scenes organized as 30 categories. Each category contains LFIs captured over the same target scene from 3 to 5 camera poses. We have selected 127 scenes\footnote{Note that we have discarded several scenes from the original MVLF dataset, over which COLMAP failed to establish correspondence. 
}, of which 109 are used for training and 18 for evaluation. We used COLMAP to estimate the camera parameters ${K, R, \tau}$ and the sparse 3D point anchors $\mathcal{P}$ based on the central views of each LF.
Since there are no ground truth disparity maps directly provided by the MVLF dataset, we generated the disparity maps using the state-of-the-art LF depth estimation method \cite{chen2018accurate} as the ground truth.

\begin{figure}[t]
\centering
\begin{subfigure}{
\centering
\includegraphics[width=1\linewidth]{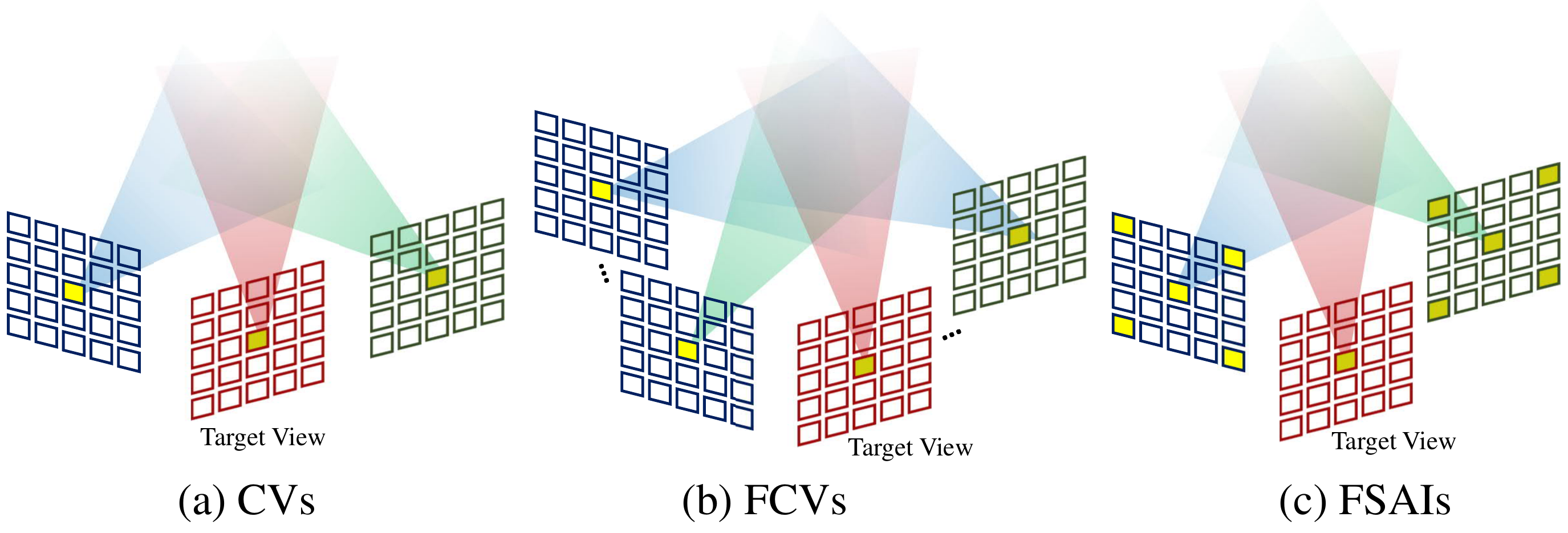}
}\end{subfigure}
\caption{Experimental set-ups for different testing cases with different SAIs from source LFIs used as source images. The selected SAIs are marked in yellow for each case.}
\label{cvs_fcvs_fsais}
\end{figure}

\textbf{\textit{Training and Implementation Details.}}
The proposed framework has been implemented with PyTorch 1.7.1. The disparity estimation model and the LF synthesis model were trained in two stages. 

For the \textit{\ul{Disparity Estimation Model}},
the \textit{Adadelta} optimizer \cite{zeiler2012adadelta} was used for training, with batch-size set to four. The learning rate was initialized to 0.01, with decay rate 0.9 starting from the tenth epoch.
Each training sample consists of one target view's color image $I_{t_0}$, ground truth disparity map $D_\text{g}$, and the pre-computed fused 3D data volume $\mathbf{U_{t_0}}\in\mathbb{R}^{376 \times 541 \times 100}$.
The feature scale number $N_\text{scale}$ for the multi-scale fusion module was set to seven,
and the number of disparity planes was set to 100.
To improve training efficiency, we have pre-calculated the target view's fused 3D data volumes $\mathbf{U}_{t_0}(x,y,d)$ off-line and used them directly during training.

For the training of the \ul{\textit{The LF Synthesis Model}}, the \textit{Adadelta} optimizer \cite{zeiler2012adadelta} was used with batch size set to one. The learning rate was initialized to 0.00001, with decay rate 0.5 starting from the second epoch.
The angular resolution of the ground truth LF in the MVLF dataset is $14\times14$, and we only render the central $7\times7$ SAIs to avoid vignetting effects. The LF synthesis model took 24 hours to train from scratch for 120 epochs on an NVIDIA Tesla V100S GPU.

\section{Evaluation and Results} \label{experiment}

In this section, we will comprehensively evaluate the efficiency of the proposed disparity estimation and the immersive LF rendering modules. We will also compare our method with several state-of-the-art novel view/LF synthesis frameworks to validate the advantages of the proposed method.

\subsection{Evaluation on Disparity Prediction}
We evaluate the disparity estimation accuracy and compare the results of our proposed method with the state-of-the-art methods, i.e., MiDaS \cite{Ranftl2020Towards}, Multi-View Stereo Network (MVSNet) \cite{yao2018mvsnet}, and  Local Light Field Fusion (LLFF) \cite{mildenhall2019llff}, based on different input configurations. Both MVSNet and LLFF can predict the target view's depth map given several source views as reference, while MiDaS works on a single image to predict the scene depth.

\textit{\textbf{For fair comparison}}, we intended to use all SAIs ($7\times 7=49$ SAIs for each LF) available from the source LFs as input for the MVSNet.
Due to the memory limit, only 11 views are allowed to be used as input.
To obtain the best results from MVSNet and LLFF, we experimented on using different input configurations, denoted as \emph{CVs}, \emph{FCVs}, and \emph{FSAIs}, respectively. As illustrated in Fig.~\ref{cvs_fcvs_fsais}(a), for \textbf{\textit{\ul{CVs}}}, the target view image, together with the central SAIs from its two neighboring source LFs are used as inputs; therefore, three images are used as inputs in total for this configuration.
For \textbf{\textit{\ul{FCVs}}} (illustrated in Fig.~\ref{cvs_fcvs_fsais}(b)), central SAIs from all LFs (4 to 6) under the same MVLF scene and the target view image are used as inputs for geometry inference.
For \textbf{\textit{\ul{FSAIs}}} (illustrated in Fig.~\ref{cvs_fcvs_fsais}(c)), the target view image, together with the central, and the four corner SAIs from its two neighboring source LFIs are used as inputs; therefore, totally 11 images are used as inputs in this case. For \textbf{\textit{\ul{CV}}}, only the target view image is used.

The depth estimation results from MiDaS, MVSNet, and LLFF, are in different scales compared with the ground truth disparity $D_{g}$. So these estimations are \textit{linearly rescaled} (including inverse operations that transforms depth to disparity values) to be aligned with the ground truth $D_{g}$. 

\begin{figure*}[h]
\centering
\begin{subfigure}{
\centering
\includegraphics[width=0.99\linewidth]{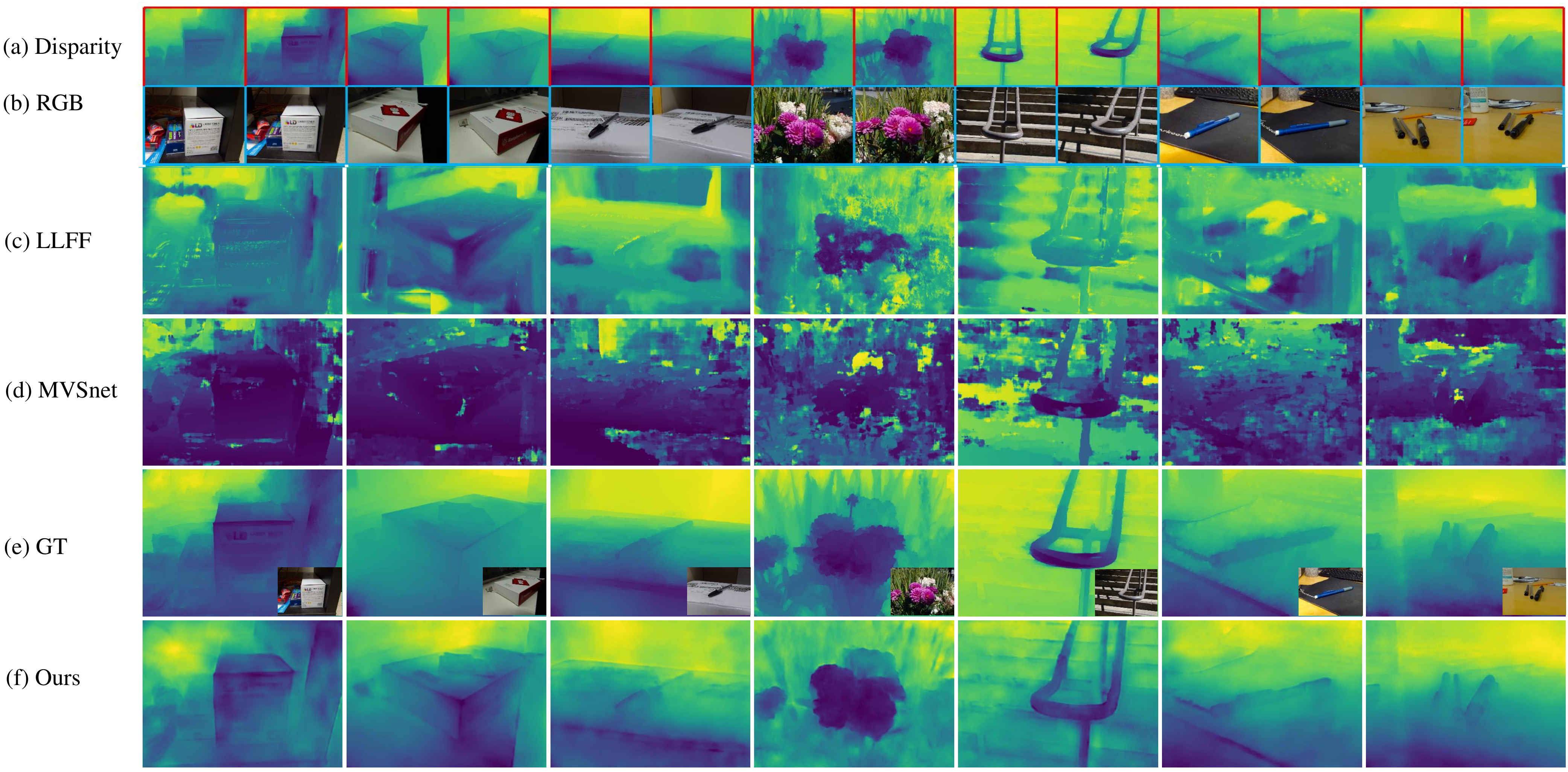}
}\end{subfigure}
\caption{Visual comparisons of predicted disparity maps between competing methods. (a) are the ground truth disparity maps from the source views (in red rectangles). (b) are the source view's RGB images (in blue rectangles). (c) are results from LLFF (CVs) \cite{mildenhall2019llff}. (d) are the results from MVSNet (CVs) \cite{yao2018mvsnet}. (e) are the ground truth disparity maps calculated by using \cite{chen2018accurate} and target view image (in the bottom-right corner). (f) are our results.}
\label{visual_comparison}\vspace{-0.4cm}
\end{figure*}

Two metrics are used to evaluate the disparity estimation quality: MSE between the prediction and the ground truth disparity, and the Percentage of Pixels with disparity estimation Errors (PPE) smaller than a given threshold (i.e., 0.05 and 0.1).
TABLE \ref{disparityevaluationresults_simple} shows the MSE and PPE results for the whole evaluation dataset, and TABLE \ref{disparityestimation} shows the evaluation detail for each scene category.
As can be seen from these results, our proposed algorithm significantly outperforms all other methods by almost \textbf{\textit{\ul{10 times}}} in the metric of MSE, and \textbf{\textit{\ul{4 times}}} in the metric of PPE. We believe this is caused by two reasons. \textit{\ul{First, local structural information from the SAIs within the same LFI (for the \textit{FSAIs} scenario) cannot be efficiently exploited in a global capture framework.}} When these locally dense SAIs are projected to the target camera's frustum, they are fused with features warped from large-baseline views. MVSNet treats all views equally and calculates feature variances to build cost volumes, which is overpowered by large baseline inputs. \textit{\ul{Second, SAIs from different LFIs have very large camera pose differences, which is too challenging for these methods to extract reliable cross-view information.}} However, our framework can fully take advantage of such local-global sampling patterns and produce impressive results.

\begin{table}[t]
\caption{Comparison of Disparity Estimation Error measured in MSE and PPE between our proposed method and MiDaS \cite{Ranftl2020Towards}, LLFF \cite{mildenhall2019llff} and MSVNet \cite{yao2018mvsnet}. Best performance is highlighted in red, second best performance in blue.}
\label{disparityevaluationresults_simple}
\centering
\begin{tabular}{llll}
\hline
Method/Metric   & MSE(px)$\downarrow$ & PPE(0.05,\%)$\uparrow$ & PPE(0.1,\%)$\uparrow$\\
\hline
\rowcolor{mygray} MiDaS (CV)  & 0.444 & 5.664 & 11.850  \\
LLFF (CVs)  &  0.394  & 4.777 & 9.678\\
\rowcolor{mygray} LLFF (FCVs) & \textcolor{blue}{0.354}  & 5.248 & 10.557  \\
LLFF (FSAIs)  & 0.447    & 4.772 & 9.789  \\
\rowcolor{mygray} MVSNet (CVs) & 0.399  & \textcolor{blue}{6.050} & \textcolor{blue}{12.139}  \\
MVSNet (FCVs) & 0.464  & 5.721 & 11.471  \\
\rowcolor{mygray} MVSNet (FSAIs) & 0.449  & 5.761 & 11.297  \\
Ours & \textcolor{red}{0.050}  & \textcolor{red}{22.878} & \textcolor{red}{41.740}  \\
\hline
\end{tabular}
\end{table}

We visually compare the estimated disparity maps between different methods in Fig. \ref{visual_comparison}. The estimation from LLFF is shown in Fig.~\ref{visual_comparison}(c). Since the LLFF framework presumes the input source views are captured from the same fronto-parallel plane in irregular grid pattern. Therefore, artifacts appear in the regions that violate such assumption -- especially for the near-camera objects that observe \textit{large, non-translational} motion between the source views. As can be seen from Fig.~\ref{visual_comparison}(d), MVSNet produces noisy outputs because the reference views from the MVLF dataset are \textit{sparse} and with \textit{large angular baseline}, which causes the generated feature volumes to show discontinuous costs along the depth planes, resulting in noisy disparity predictions. In addition, noisy estimations happen not only in texture-less or occlusion ambiguous regions, but in general areas with \textit{larger} angular parallaxes. This shows that the MVSNet framework is generally unable to deal with sparse and large angular baseline source view inputs.
Fig.~\ref{visual_comparison}(f) shows the results from our proposed method. The estimations adhere to the ground truth much better than the competing methods.

\begin{figure}[t]
\centering
\begin{subfigure}{
\centering
\includegraphics[width=0.98\linewidth]{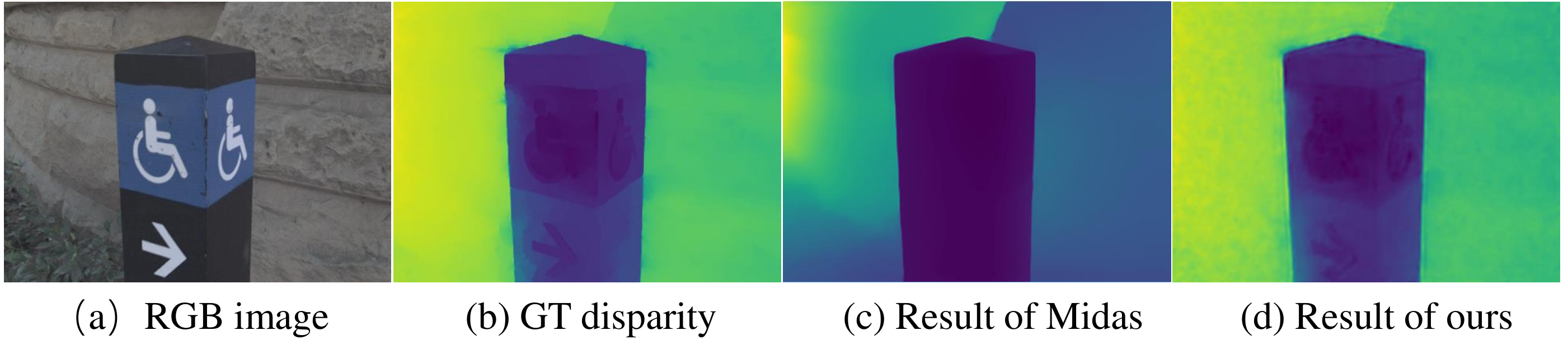}
}\end{subfigure}
\caption{Visual Comparison with MiDaS. MiDaS can generate visually pleasant disparity result as in (c). But the disparity scale, and the structural details are generally incorrect.}
\label{compare_midas}\vspace{-0.3cm}
\end{figure}

We also visually compare our results with MiDaS \cite{Ranftl2020Towards} in Fig. \ref{compare_midas}, which generates visually pleasant disparity result based on a single input RGB image. But the relative disparity scale, and the structural details within the scene are generally incorrect. This is due to its monocular pipeline, which lacks reliable geometrical clues compared with those exploited in the multi-view frameworks.

\subsection{Evaluation on Light Field Rendering Quality}

We evaluate the quality of the final synthesized LFIs both quantitatively and qualitatively and compare with two state-of-the-art LF synthesis methods with relevant context, i.e., LLFF, which takes a group of local captures and synthesizes a local immersive LF at a given camera position via fusion of multiple multi-plane image (MPI) representations; and SynSin \cite{wiles2020synsin}, which takes a single RGB image as input, based on which a depth map is estimated to form a point cloud representation. Point-based rendering techniques are then employed to render novel views at different angles. For both LLFF and SynSin, an immersive LF with the target image as central SAI can be synthesized by specifying the camera's extrinsic parameters for each SAI, which we estimate based on the ground truth LFIs from the testing dataset using COLMAP.

\begin{figure*}[t]
\centering
\begin{subfigure}{
\centering
\includegraphics[width=0.98\linewidth]{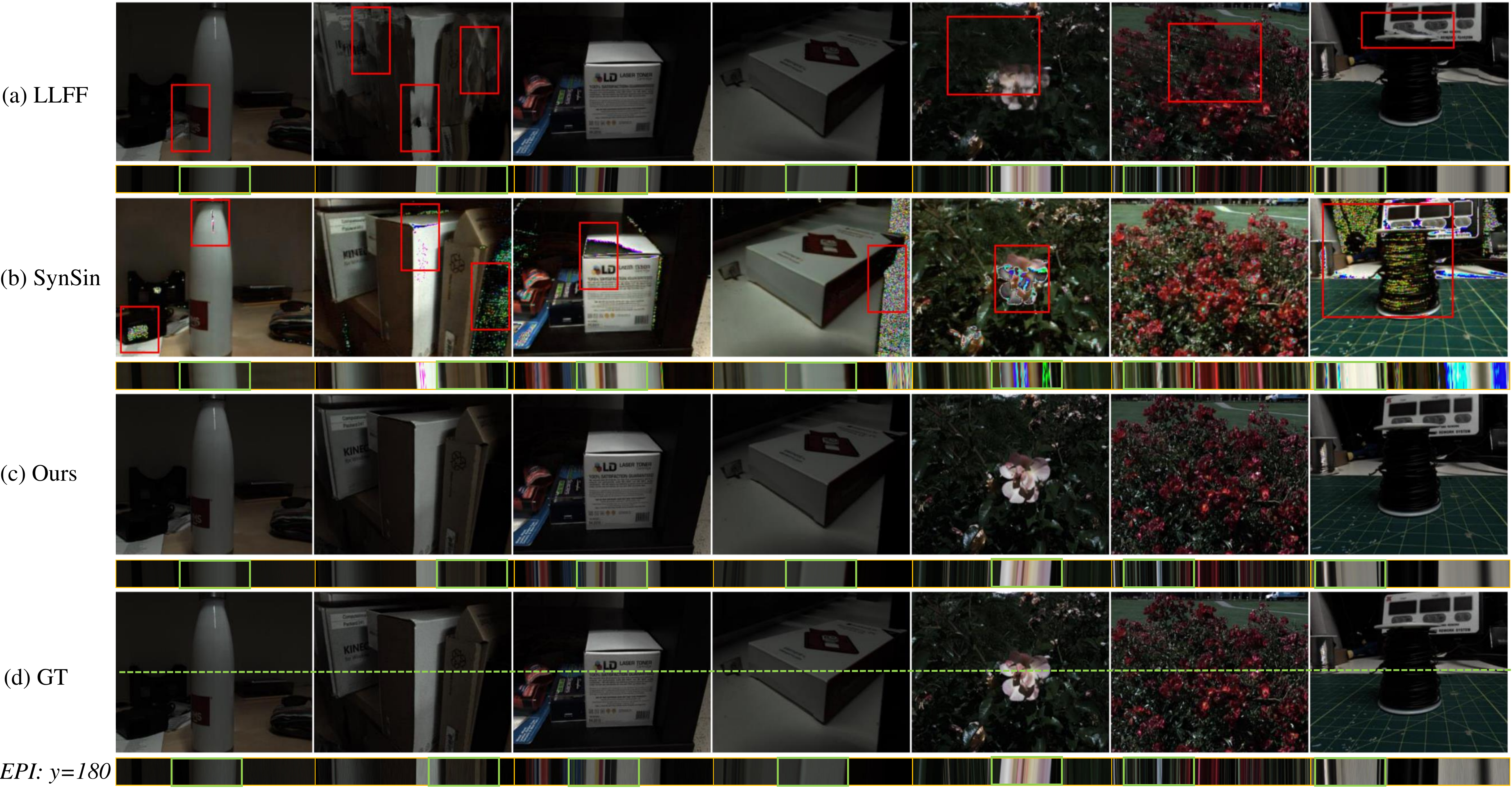}
}\end{subfigure}
\caption{Light Field synthesis visual comparison. We compare the results of our method in row (c) with those of other competing view synthesis methods, i.e., LLFF(CVs)\cite{mildenhall2019llff} in row (a) and SynSin(CV) \cite{wiles2020synsin} in row (b). The ground truth SAI is in row (d). We show the top-left SAI from the $7\times7$ LFIs and their respective EPIs sampled at $y$=180. LLFF shows poor performance on thin structures and has aliasing effect around the edges. SynSin suffers from noisy pixels in bright and dark region. Our results align much better to the ground truth both for the SAIs' visual quality, and for the linear structure preserved in the EPIs.}
\label{lfsyn}\vspace{-0.4cm}
\end{figure*}

We quantitatively evaluate the quality of the synthesized LFIs using two metrics, i.e., Peak-Signal Noise Ratio (PSNR) and Structural Similarity (SSIM), and the results are shown in TABLE \ref{lightfieldsynthesis}. As can be seen, our method significantly outperforms other competing methods quantitatively by \textbf{\textit{\ul{8 dB}}} against LLFF, and \textbf{\textit{\ul{20 dB}}} against SynSin.
Fig. \ref{lfsyn} gives a qualitative presentation of the synthesized LF. We can observe that LLFF shows poor performance over thin structures and has an aliasing effect around the object edges. SynSin suffers from noisy pixels in bright and dark regions and shows very obvious structural distortions at off-set views. Our results align much better with the ground truth both for the visual quality of the SAIs and for the linear structure preserved in the EPIs, indicating that our method is able to render both spatial and angular immersive details truthfully, contributed by the high-quality disparity field estimation as well as the deep re-regularization modules.

\subsection{Ablation study}

\textbf{\textit{Disparity Estimation Module.}}
We carried out ablation study to evaluate the contribution of the Frustum Voxel Filtering Module and the Attention-Guided Multi-scale Residual Fusion Module for the disparity estimation.

Fig.~\ref{ablation_study} qualitatively demonstrates the contributions from these modules. We can observe significant improvements in \textbf{\textit{completeness}} and \textbf{\textit{semantic correctness}} in the refined disparity maps $\hat{D}_{t_0}$ as compared with the preliminary, and initial disparity estimations $D_{t_0}$, $D_\text{init}$.

To quantitatively validate the respective contributions of each sub-module of the disparity estimation model, networks trained with different combinations of sub-modules are tested, and the results are presented in TABLE~\ref{ablationstudy}. As can be seen, without the attention units in the Attention-Guided Multi-scale Residual Fusion Module (\textit{\textbf{Ours-wo-Atten}}), we observe only slight performance degradation in the metric of MSE; however, the impact is more significant on PPE. This indicates that the attention mechanism is efficient for detecting \textit{\textbf{regions with larger errors}} (above the thresholds), which helps the network to focus on improving via residual fusion. When our model is without RGB image as guide (\textit{\textbf{Ours-wo-RGB}}), we observe a sharp degradation in the metric of MSE. RGB is important for propagating textural and semantic information to the disparity.

\begin{table}[t]
\caption{Ablation study on disparity estimation. The most import factor (measured by greatest performance degradation when absent) is highlighted in red.
}\label{ablationstudy}
\centering
\label{ablation}
\renewcommand\tabcolsep{3.0pt}
\begin{tabular}{lllllllllllllllll}
\hline
Method & Dataset Average \\
Metric  & MSE(px)$\downarrow$ &PPE(0.05, \%)$\uparrow$ & PPE(0.1, \%)$\uparrow$ \\
\hline
\rowcolor{mygray}Ours-wo-Atten   & 0.063 & \textcolor{red}{18.826} & \textcolor{red}{35.528} \\
Ours-wo-RGB   & \textcolor{red}{0.108}  & 20.699 & 35.872\\
\rowcolor{mygray} Ours  & 0.050  & 22.878 & 41.740 \\
\hline
\end{tabular}
\end{table}

\begin{table}[t]
\caption{ Ablation study on LF synthesis.}\label{ablationstudyLFsyn}
\centering
\label{ablation}
\renewcommand\tabcolsep{3.0pt}
\begin{tabular}{lllllllllllllllll}
\hline
Method & Dataset Average \\
Metric  &  PSNR$\uparrow$ & SSIM$\uparrow$ \\
\hline
Ours-wo-disparity-field &\textcolor{red}{30.073} & \textcolor{red}{0.884} \\
Ours  & 33.351 & 0.915\\
\hline
\end{tabular}\vspace{-0.4cm}
\end{table}

\textbf{\textit{Light Field Synthesis Module.}}
The Spatial-Angular Convolution Module has been widely adopted and proven to be useful in various LF synthesis frameworks \cite{8561240, article, yeung2018fast}. We only validate the functionality of disparity field estimation network by replacing it with directly copied central view's disparity $\hat{D}_{t_0}$. The evaluations are conducted on LF rendering quality by calculating the PSNR and SSIM against ground truth LF $\mathbf{L}_\text{gt}$. The results are shown in TABLE~\ref{ablationstudyLFsyn}. We can see that a performance gain of \textbf{\textit{\ul{3.278 dB}}} is achieved by the disparity field synthesis module.

\begin{table*}[htbp]
\caption{ The MSE of disparity estimation in test scenes, best performance is highlighted in red, second best performance is highlighted in blue. $\downarrow$ denotes the smaller, the better.} \label{disparityestimation}
\centering
\begin{tabular}{llllllllllll}
\hline
Scene Category   & Bottles & Boxes & Cacti & Flowers & Leaves & Misc & Pens & Average \\
Method & MSE(px)$\downarrow$ & MSE(px) & MSE(px) & MSE(px) & MSE(px) & MSE(px) & MSE(px) & MSE(px) \\
\hline
\rowcolor{mygray}MiDaS \cite{Ranftl2020Towards} (CV)  & \textcolor{blue}{0.217} & 0.402 & \textcolor{blue}{0.127} & \textcolor{blue}{0.246} & \textcolor{blue}{0.284} & 0.554 & 1.086 & 0.444\\
LLFF\cite{mildenhall2019llff} (CVs)  & 0.234 & \textcolor{blue}{0.377} & 0.400 & 0.339 & 0.390 & 0.385 & 0.602 &   0.394 \\
\rowcolor{mygray}LLFF\cite{mildenhall2019llff} (FCVs)  & 0.225 & 0.427 & 0.252 & 0.343 & 0.419 & \textcolor{blue}{0.223}  & \textcolor{blue}{0.355} & \textcolor{blue}{0.354}\\
LLFF\cite{mildenhall2019llff} (FSAIs)  & 0.225 & 0.506 & 0.455 & 0.349 & 0.451 & 0.478 & 0.651 & 0.447  \\
\rowcolor{mygray}MVSNet\cite{yao2018mvsnet} (CVs)  & 0.342 & 0.382 &0.360 & 0.370 & 0.450 & 0.416 & 0.432 & 0.399 \\
MVSNet\cite{yao2018mvsnet} (FCVs)  & 0.556 & 0.473 & 0.343 & 0.451 & 0.434 & 0.472 & 0.510 & 0.464\\
\rowcolor{mygray}MVSNet\cite{yao2018mvsnet} (FSAIs) & 0.427 & 0.441 & 0.446  & 0.372 & 0.433 & 0.509 & 0.569 & 0.449 \\
Ours & \textcolor{red}{0.017} & \textcolor{red}{0.053}& \textcolor{red}{0.013} & \textcolor{red}{0.031} & \textcolor{red}{0.054} & \textcolor{red}{0.035} & \textcolor{red}{0.106} & \textcolor{red}{0.050} \\
\hline
\end{tabular}
\label{disparityevaluationresults}
\end{table*}

\begin{table*}[htbp]
\caption{ LF synthesis quality evaluation. Best performance highlighted in red, second best in blue. $\uparrow$ denotes the larger, the better.}\label{lightfieldsynthesis}
\centering
\renewcommand\tabcolsep{3.0pt}
\begin{tabular}{lllllllllllllllll}
\hline
Scene Category   & Bottles && Boxes && Cacti && Flowers && Leaves && Misc && Pens && Average \\
Method & PSNR$\uparrow$& SSIM$\uparrow$ & PSNR& SSIM & PSNR& SSIM & PSNR& SSIM & PSNR& SSIM & PSNR & SSIM & PSNR & SSIM & PSNR & SSIM  \\
\hline

\rowcolor{mygray}LLFF\cite{mildenhall2019llff} (CVs)  & \textcolor{blue}{31.279}  & {0.870} & \textcolor{blue}{30.356} & \textcolor{blue}{0.850} & \textcolor{blue}{20.101} & \textcolor{blue}{0.568} & \textcolor{blue}{22.411} & \textcolor{blue}{0.569} & \textcolor{blue}{22.806} & \textcolor{blue}{0.716} & \textcolor{blue}{24.449} & \textcolor{blue}{0.749} & \textcolor{blue}{30.273} & \textcolor{blue}{0.855} & \textcolor{blue}{25.702} & \textcolor{blue}{0.725} \\
LLFF\cite{mildenhall2019llff} (FCVs)  & 30.608 & \textcolor{blue}{0.878} & 24.110  & 0.728 & 18.259 & 0.446 & 19.480 &  0.394 &  19.393 & 0.509 & 15.496 & 0.439 & 11.363 & 0.435 & 18.992 & 0.511  \\
\rowcolor{mygray}LLFF\cite{mildenhall2019llff} (FSAIs)  & 29.649  & 0.852 & 28.782 & 0.807 & 18.597 & 0.474 & 20.602 & 0.496 & 18.351 & 0.432 & 23.322 & 0.724 & 29.712 & 0.861 & 23.802 & 0.642  \\
SynSin\cite{wiles2020synsin} (CV) & 16.722 & 0.732 & 14.567 & 0.656 & 6.152 & 0.017 & 14.639 & 0.419 & 13.242 & 0.325 & 9.946 & 0.230 & 15.433 & 0.696 & 13.649 & 0.463\\

\rowcolor{mygray}Ours   & \textcolor{red}{39.060} & \textcolor{red}{0.952} & \textcolor{red}{36.740} & \textcolor{red}{0.937} & \textcolor{red}{32.631} & \textcolor{red}{0.965} & \textcolor{red}{30.570} & \textcolor{red}{0.896} & \textcolor{red}{28.124} & \textcolor{red}{0.854} & \textcolor{red}{33.495} & \textcolor{red}{0.944} & \textcolor{red}{38.064} & \textcolor{red}{0.938} & \textcolor{red}{33.351 }& \textcolor{red}{0.915} \\
\hline
\end{tabular} \vspace{-0.4cm}
\end{table*}

\section{Concluding Remarks} \label{concludingremarks}

We have proposed a novel view/LF synthesis framework which robustly fuses and transfers scene geometry from large baseline LF captures.
As far as we know, this is one of the first attempts for an integrated modeling scheme which transfers the requirement for a \textit{globally-dense sampling} to a \textit{sparse set of locally-dense sampling} (in the forms of distributed LF captures). In this sense, our work can facilitate cheap and convenient capture of target scenes.
We have proposed a novel scale-consistent frustum volume rescaling algorithm which enables fusion of distributed, heterogeneous geometry embedding to be globally consistent. We have proposed novel learning-based processing modules which comprehensively regularize noisy observations from heterogeneous captures and fuse these complementary features for high-quality rendering of both disparity maps and novel LFIs.
Both quantitative and qualitative experiments show that our proposed method produces precise and high-quality disparity estimation and LFIs at the target view. Our method outperforms the alternative state-of-the-art methods significantly under similar capture configurations.
For future work, we plan to further explore the complementary information from different source DPVs and investigate efficient modeling mechanisms for occlusion-aware content completion and view-depend effect rendering.

\bibliographystyle{IEEEbib}
\bibliography{mybib}

\end{document}


\title{Scale-Consistent Fusion: from Heterogeneous Local Sampling to Global Immersive Rendering\\
Supplementary Material}
\maketitle

\section{Introduction}
This document serves as a supplementary material to the manuscript: \textit{Scale-Consistent Fusion: from Heterogeneous Local Sampling to Global Immersive Rendering}. Sec. \ref{training} illustrates the complete name list of training scenes used in our training process from Stanford Lytro Multi-view Light Field Dataset \cite{dansereau2019liff}. Sec. \ref{testing} shows the complete test results on all test scenes, including the results of MVSNet \cite{yao2018mvsnet}, LLFF \cite{mildenhall2019llff}, MiDaS \cite{Ranftl2020Towards} and Ours.

Please note that a video file \textit{Supplementary\_video.mp4} which shows the volumes fusing results and the final synthesized novel views is also provided as a separate document.

\section{List of Training Scenes}\label{training}
The complete list of training scenes used are batteries\_0, cables\_12, flowers\_50, misc\_27, batteries\_1, cables\_21, flowers\_54, misc\_8, bikes\_11, cables\_25,  flowers\_79, pens\_and\_pencils\_0, books\_0, cables\_26, flowers\_87, pens\_and\_pencils\_1, books\_2, cables\_33, glasses\_1, pens\_and\_pencils\_11, bottles\_0, cables\_34, glasses\_6, pens\_and\_pencils\_20, bottles\_2, cables\_35, glue\_3, phones\_0, bottles\_4, cables\_36, glue\_4, phones\_1, boxes\_0, cables\_39, keyboards\_5, hones\_3, boxes\_1, cables\_7, keyboards\_6, phones\_4, boxes\_10, cables\_8, leaves\_111, screws\_2, boxes\_12, cacti\_1, leaves\_112, screws\_3, boxes\_13, cacti\_5, leaves\_117, screws\_6, boxes\_16, chairs\_2, eaves\_132, signs\_12, boxes\_17, coins\_5, leaves\_15, signs\_29, boxes\_20, coins\_6, leaves\_2, succulents\_11, boxes\_21, cups\_1, leaves\_20, succulents\_7, boxes\_22, cups\_10, leaves\_46, tools\_10, boxes\_23, cups\_11, leaves\_59, tools\_2, boxes\_25, cups\_4, leaves\_69, tools\_6, boxes\_26, cups\_6, misc\_0, tools\_7, boxes\_27, flowers\_104, misc\_1, tools\_8, boxes\_29, lowers\_113, misc\_12, tools\_9, boxes\_4, flowers\_136, misc\_15, trees\_14, boxes\_9, flowers\_15, misc\_2, trees\_55, buildings\_1, flowers\_166, misc\_22, buildings\_61, flowers\_172, misc\_23, cables\_1, flowers\_187, misc\_26.

\section{Test Results} \label{testing}

The complete results of disparity evaluation on all test scenes are shown in TABLE \ref{CompleteDispMVS}, \ref{CompleteDispLLFF} and \ref{CompleteDispMidas}. 

\begin{table*}[h]
\caption{The complete disparity estimation quality evaluation results of MVSNet on test scenes. $\uparrow$ denotes the larger, the better. $\downarrow$ denotes the smaller, the better.}
\centering
\renewcommand\tabcolsep{3pt} 

\label{CompleteDispMVS}
\begin{tabular}{lllllllllllllllllll}
\hline
Method & \multicolumn{3}{c}{MVSNet \cite{yao2018mvsnet} (CVs)} & \multicolumn{3}{c}{MVSNet \cite{yao2018mvsnet} (FCVs)} & \multicolumn{3}{c}{MVSNet \cite{yao2018mvsnet} (FSAIs)} \\ 
Scene & MSE (px) $\downarrow$ & PPE (0.05, \%) $\uparrow$ & PPE (0.1, \%) $\uparrow$ & MSE (px) & PPE (0.05, \%) & PPE (0.1, \%) & MSE (px) & PPE (0.05\%) & PPE (0.1\%)  \\ 
\hline
\rowcolor{mygray} bottles\_6 & 0.343 & 9.228 &  13.896 & 0.556 & 6.479 &9.195 & 0.427  & 6.854 & 11.072\\ 
boxes\_19 & 0.414 & 4.686 & 9.368 &0.528  & 3.969& 7.481&  0.486 & 4.931 & 9.945\\
\rowcolor{mygray}boxes\_6 &0.308 & 7.840 & 13.956 & 0.306 & 7.010& 14.048 & 0.374 & 5.789 &11.848\\ 
boxes\_7 & 0.426 & 3.676 & 7.951 & 0.587 & 2.944 & 6.314 & 0.464  & 2.147&5.690\\ 
\rowcolor{mygray}cacti\_11 & 0.361 & 5.998 & 12.733 & 0.343 & 8.078 & 14.686 & 0.446 & 7.236 & 13.399\\ 
flowers\_119 & 0.285 & 9.642 & 20.131 & 0.416 & 5.664 &12.166 & 0.257 & 10.454 & 20.442\\
\rowcolor{mygray}flowers\_121 & 0.307 & 5.286 & 10.966& 0.344 & 4.470 & 9.150 & 0.403 &4.278&8.496\\
flowers\_150 & 0.366 & 5.288 & 10.545 & 0.436 & 4.211&8.457 & 0.458 & 3.233 & 6.962\\
\rowcolor{mygray}flowers\_182 & 0.426 &6.204 &14.619& 0.613 &7.293& 17.010 & 0.312 & 11.160 & 19.842\\
flowers\_31 & 0.467 & 4.593 & 8.810& 0.450 &4.487 & 8.600 & 0.433 & 4.615 & 9.196\\
\rowcolor{mygray}leaves\_48 & 0.666 & 4.182 &  8.562 &  0.686& 4.685 & 9.397 &  0.704 &3.066& 8.635\\
leaves\_55 & 0.215 & 9.412 & 19.234&  0.212 & 11.097 & 20.351 & 0.249 & 7.426 & 15.713\\
\rowcolor{mygray}leaves\_75 &0.472 &3.917 &9.013& 0.405 & 5.139 & 11.933 &0.346  &5.932&11.573\\
misc\_17 & 0.295 & 9.775 & 21.123 & 0.348 & 9.646& 20.198 & 0.463 & 10.658&19.128\\
\rowcolor{mygray}misc\_21 & 0.539 & 5.040 & 9.244& 0.598 & 3.814 & 8.025 &0.557   & 4.568&8.640 \\
pens\_10 & 0.372 & 4.355 & 8.693& 0.316 & 6.648 & 14.596 &  0.561 & 3.497 & 6.973\\
\rowcolor{mygray}pens\_13 & 0.509 & 4.073 & 8.724& 0.537 &4.187 & 8.720& 0.553  & 5.535 & 11.091\\
pens\_14 & 0.417 & 5.720 & 10.944& 0.678 & 3.162 & 6.161 &  0.596 & 2.331&4.704\\ \hline
Average & 0.399 & 6.050 &12.139&  0.464 & 5.721 & 11.471 & 0.449&5.761&11.297\\ \hline

\end{tabular} \vspace{-0.4cm}
\end{table*}

\begin{table*}[h]
\caption{The complete disparity quality evaluation results of LLFF on test scenes. $\uparrow$ denotes the larger, the better. $-$ denotes the failure of reconstruction, $\downarrow$ denotes the smaller, the better.}
\centering
\renewcommand\tabcolsep{3pt} 

\label{CompleteDispLLFF}
\begin{tabular}{lllllllllll}
\hline
Method & \multicolumn{3}{c}{LLFF \cite{mildenhall2019llff} (CVs)}
& \multicolumn{3}{c}{LLFF \cite{mildenhall2019llff} (FCVs)}
& \multicolumn{3}{c}{LLFF \cite{mildenhall2019llff} (FSAIs)} \\ 
Scene & MSE (px) $\downarrow$ & PPE(0.05, \%) $\uparrow$ & PPE(0.1, \%) $\uparrow$ & MSE (px) & PPE (0.05, \%) & PPE (0.1, \%) & MSE (px) & PPE (0.05, \%) & PPE (0.1, \%) \\ 
\hline
\rowcolor{mygray} bottles\_6 & 0.234 & 3.836 & 8.163  &0.225  & 4.242 & 9.706 &0.226   & 5.297 & 12.838\\ 
boxes\_19 & 0.515 & 6.987 & 14.652  & 0.687 &4.438 & 8.808 & 0.812  & 5.816 &10.693 \\
\rowcolor{mygray}boxes\_6 & 0.213 & 4.092 & 8.504  & 0.208 & 4.873 & 10.189 &  0.273 & 6.894 &14.219 \\ 
boxes\_7 & 0.404 & 5.811 &  10.799 & 0.389 & 6.975 & 12.218 & 0.433  & 5.005 & 10.037\\ 
\rowcolor{mygray}cacti\_11 & 0.267 & 4.366 & 9.122  & 0.253 &4.254  & 9.377 &  0.283 & 3.456 & 7.912\\ 
flowers\_119 & 0.554 & 3.369  & 7.542  &0.522 & 5.748 & 11.750 &  0.507 & 2.719 &5.795 \\
\rowcolor{mygray}flowers\_121 & 0.280 & 4.687 & 9.844  & 0.295& 4.533 & 9.259 & 0.325   &4.167  & 8.745\\
flowers\_150 & 0.270 & 4.155 &  9.042 & 0.275 & 5.630 & 11.534 &  0.311 & 4.129 & 7.844\\
\rowcolor{mygray}flowers\_182 & 0.352 & 9.481 &  18.036 & 0.374 &9.618 &  18.978 & 0.364 & 10.386 & 19.246\\
flowers\_31 & 0.243 & 3.725  & 7.481  & 0.253 &3.572 & 7.452 & 0.240  & 3.555 &8.021 \\
\rowcolor{mygray}leaves\_48 & 0.377 & 7.603 &  14.398 & 0.345&7.190 & 14.581 &  0.232 & 6.839 & 14.971\\
leaves\_55 & 0.477 & 3.254 & 6.454  &0.627 &1.478 & 3.001 &  0.845 & 1.031 &2.129 \\
\rowcolor{mygray}leaves\_75 & 0.319 & 5.341  &  10.821 &0.286 & 4.689 &9.279& 0.278  & 8.379 & 16.586\\
misc\_17 & 0.213  &4.789 &  10.314 & 0.224 &6.234 & 11.678 &  0.237 & 4.231 & 8.567\\
\rowcolor{mygray}misc\_21 & 0.558 & 2.752 & 6.017  &- &- &-&  0.720 &  3.730&7.800 \\
pens\_10 & 0.675 & 3.737 &  7.353 & -&- &-& 0.768  & 3.358 &6.564\\
\rowcolor{mygray}pens\_13 & 0.739 & 3.980 &  7.714 & -&- &-& 0.727  & 4.813 &9.839 \\
pens\_14 & 0.395 & 4.036 &  7.954 &- &- &-& 0.459  &  2.102 & 4.413\\ \hline
Average & 0.394 & 4.777 & 9.678  & 0.354 & 5.248& 10.557 &0.447 &4.772&9.789 \\ \hline
\end{tabular} \vspace{-0.4cm}
\end{table*}

\begin{table*}[h]
\caption{The complete disparity quality evaluation of MiDaS and Ours on test scenes. $\uparrow$ denotes the larger, the better. $\downarrow$ denotes the smaller, the better.}
\centering
\renewcommand\tabcolsep{3pt} 

\label{CompleteDispMidas}
\begin{tabular}{lllllllllllllllllll}
\hline
Method & \multicolumn{3}{c}{MiDaS \cite{Ranftl2020Towards} (CV)} & \multicolumn{3}{c}{Ours (CVs)}\\ 
Scene & MSE (px) $\downarrow$ & PPE (0.05, \%) $\uparrow$ & PPE (0.1, \%) $\uparrow$ & MSE (px) & PPE (0.05, \%) & PPE (0.1, \%)  \\ 
\hline
\rowcolor{mygray} bottles\_6 & 0.217 & 2.676 & 5.633  & 0.018 & 28.975 & 55.278\\ 
boxes\_19 & 0.873 & 4.466 & 8.728  & 0.106 & 11.060 & 21.446\\
\rowcolor{mygray}boxes\_6 & 0.232  & 1.154 & 2.918  & 0.027 & 23.938 & 45.311  \\ 
boxes\_7 & 0.103 & 4.108 & 10.173  & 0.027 & 17.298 & 17.298 \\ 
\rowcolor{mygray}cacti\_11 & 0.128 & 9.818 & 16.813  & 0.014 & 35.609 &60.172 \\ 
flowers\_119 & 0.720 & 6.246 & 11.550  & 0.079 & 17.026 & 32.858 \\
\rowcolor{mygray}flowers\_121 & 0.143 & 11.841 &  19.754  & 0.023  & 24.758 & 48.221\\
flowers\_150 & 0.119 & 9.127 & 19.434  & 0.024 & 23.306 & 44.178\\
\rowcolor{mygray}flowers\_182 & 0.232 & 4.201 & 9.211  & 0.026 & 27.317 & 47.452 \\
flowers\_31 & 0.018 & 17.234 & 38.683  & 0.006 & 57.222 &  81.423\\
\rowcolor{mygray}leaves\_48  & 0.244 & 4.202& 8.766 & 0.068 & 13.718 & 25.988\\
leaves\_55 & 0.459 & 4.129 &  7.955 & 0.048 & 13.122  & 27.459\\
\rowcolor{mygray}leaves\_75 &  0.150 & 2.578 &  6.011 & 0.048 & 16.918 & 34.091\\
misc\_17 & 0.043 &10.545  & 28.838  & 0.009 & 37.986 & 67.675 \\
\rowcolor{mygray}misc\_21 &  1.066 & 1.388 &  2.968 &  0.062& 17.209 &33.580 \\
pens\_10 & 1.725  & 2.722 & 5.143  & 0.175 & 9.101 & 18.126  \\
\rowcolor{mygray}pens\_13 &  0.737 &  3.258 &  6.105 & 0.103 & 13.467 &  26.517 \\
pens\_14 & 0.798 & 2.259 &  4.629 & 0.041  & 23.776 & 44.642 \\ \hline
Average & 0.444  & 5.664 & 11.850  & 0.050 & 22.878 & 41.740 \\ \hline
\end{tabular} \vspace{-0.4cm}
\end{table*}

The complete results of LF synthesis quality evaluation on all test scenes are shown in TABLE \ref{CompleteLFsyn}.

\begin{table*}[h]
\caption{The complete LF synthesis quality evaluation on test scenes. $\uparrow$ denotes the larger, the better.}
\centering
\renewcommand\tabcolsep{5pt} 

\label{CompleteLFsyn}
\begin{tabular}{lllllllllll}
\hline
Method & \multicolumn{2}{c}{SynSin \cite{wiles2020synsin} (CV)} & \multicolumn{2}{c}{LLFF \cite{mildenhall2019llff} (CVs)} & \multicolumn{2}{c}{LLFF \cite{mildenhall2019llff} (FCVs)}& \multicolumn{2}{c}{LLFF \cite{mildenhall2019llff} (FSAIs)} & \multicolumn{2}{c}{Ours}  \\ 
Scene & PSNR $\uparrow$ & SSIM $\uparrow$ & PSNR & SSIM & PSNR & SSIM  & PSNR & SSIM & PSNR & SSIM\\ 
\hline
\rowcolor{mygray} bottles\_6 & 16.722 & 0.732  & 31.280 & 0.870 & 30.608 & 0.879  & 29.649 & 0.853  & 39.060 & 0.952\\ 
boxes\_19 & 13.044 & 0.6771 & 33.210 & 0.901 & 27.532 & 0.821 & 34.726 & 0.919 & 39.915 & 0.969\\
\rowcolor{mygray}boxes\_6 & 15.301 & 0.608 & 32.123 & 0.835 & 24.163 & 0.708 & 23.830 & 0.652 & 38.308 & 0.919 \\ 
boxes\_7 & 15.356 & 0.682  & 25.735 & 0.817 & 20.637 & 0.657 & 27.793 & 0.851 & 31.998 & 0.925 \\ 
\rowcolor{mygray}cacti\_11 & 6.152 & 0.017 & 20.101 & 0.569 & 18.259 & 0.447 & 18.598 & 0.474  & 32.631 & 0.966 \\ 
flowers\_119 & 17.859 & 0.558 & 21.348 & 0.508 & 20.292 & 0.474 & 19.919 & 0.491 & 27.757 & 0.825 \\
\rowcolor{mygray}flowers\_121 & 6.187 & 0.014 & 24.663 & 0.724 & 19.797 & 0.422 & 21.127 & 0.578 & 34.946 & 0.964 \\
flowers\_150 & 16.034 & 0.464 & 20.823 & 0.474 & 18.715 & 0.272 & 19.718 & 0.351 & 28.384 & 0.895\\
\rowcolor{mygray}flowers\_182 & 15.935 & 0.573 & 27.308 & 0.779  & 21.886 & 0.472 & 24.304 & 0.713  & 34.625 & 0.947 \\
flowers\_31 & 17.179 & 0.489 & 17.918 & 0.363 & 16.713 & 0.333 & 17.943 & 0.351 & 27.141 & 0.851 \\
\rowcolor{mygray}leaves\_48 & 16.809 & 0.527 & 23.307 & 0.749 & 17.875 & 0.390 & 13.845 & 0.074 & 28.039 & 0.895 \\
leaves\_55 & 16.288 & 0.413 & 22.961 & 0.667 & 20.907 & 0.519  & 20.846 & 0.573 & 23.570 & 0.699 \\
\rowcolor{mygray}leaves\_75 & 6.628 & 0.036 & 22.152 & 0.732 & 19.399 & 0.619 & 20.364 & 0.652 & 32.764 & 0.970 \\
misc\_17 & 6.691 & 0.022 & 25.021 & 0.811 & 20.518 & 0.636  & 22.836 & 0.744  & 35.499 & 0.970\\
\rowcolor{mygray}misc\_21 & 13.201 & 0.439 & 23.879 & 0.688 & 10.474 & 0.242 & 23.809 & 0.705  & 31.493 & 0.919 \\
pens\_10 & 15.817 & 0.710 & 32.928 & 0.902  & 10.312 & 0.526 & 32.450 & 0.895  & 40.535 & 0.965\\
\rowcolor{mygray}pens\_13 & 15.387 & 0.623 & 26.572 & 0.794 & 13.080 & 0.333  & 23.781 & 0.785 & 28.684 & 0.831\\
pens\_14 & 15.097 & 0.754 & 31.320 & 0.870  & 10.697 & 0.448 & 32.907 & 0.905  & 38.656 & 0.961\\ \hline
Average & 13.649 &0.463 & 25.702 & 0.725 & 18.992 & 0.511 &23.802 & 0.642 & 33.351 &0.915\\ \hline

\end{tabular} \vspace{-0.4cm}
\end{table*}

\newpage
\bibliographystyle{IEEEbib}
\bibliography{mybib}